\documentclass{article} 
\usepackage[final]{colm2026_conference}

 \usepackage{comment}

\usepackage{microtype}
\usepackage{hyperref}
\usepackage{url}
\usepackage{booktabs}
\usepackage{amsmath}
\usepackage{amssymb}
\usepackage{graphicx}
\usepackage{enumitem}
\setlist[itemize]{noitemsep, topsep=0pt}

\usepackage{multirow}

\usepackage[acronym]{glossaries}
\newacronym{cmp}{CMP}{Cross-Model Perplexity}
\newacronym{cme}{CME}{Cross-Model Entropy}

\makeglossaries

\usepackage{titlesec}
\titlespacing*{\section}{0pt}{1pt}{1pt}
\titlespacing*{\subsection}{0pt}{1pt}{1pt}
\titlespacing*{\paragraph}{0pt}{1pt}{0.5em}

\usepackage{lineno}

\definecolor{darkblue}{rgb}{0, 0, 0.5}
\hypersetup{colorlinks=true, citecolor=darkblue, linkcolor=darkblue, urlcolor=darkblue}

\makeatletter
\def\section{\@startsection{section}{1}{\z@}%
  {-2.0ex plus -0.5ex minus -.2ex}%
  {0.6ex plus 0.2ex minus 0.1ex}%
  {\large\bf\raggedright}}

\def\subsection{\@startsection{subsection}{2}{\z@}%
  {-0.8ex plus -0.2ex minus -.1ex}%
  {0.4ex plus 0.1ex}%
  {\normalsize\bf\raggedright}}
\makeatother

\title{Cross-Model Disagreement as a Label-Free Correctness Signal}

\author{Matt Gorbett \\
Independent Researcher \\
\texttt{matthewgorbett@gmail.com} \\
\And
Suman Jana \\
Department of Computer Science \\
Columbia University \\
New York, NY 10027, USA \\
\texttt{sj2536@columbia.edu}
}

\begin{document}

\titlespacing*{\section}{0pt}{8pt plus 2pt minus 2pt}{4pt plus 1pt minus 1pt}
\titlespacing*{\subsection}{0pt}{6pt plus 2pt minus 2pt}{3pt plus 1pt minus 1pt}

\ifcolmsubmission
\linenumbers
\fi

\maketitle

\vspace{-2em}

\begin{abstract}
Detecting when a language model is wrong without ground truth labels
is a fundamental challenge for safe deployment. Existing approaches
rely on a model's own uncertainty, such as token entropy or confidence
scores, but these signals fail critically on the most dangerous failure
mode: confident errors, where a model is wrong but certain. In this work we introduce \textit{cross-model disagreement} as a correctness indicator — a simple, training-free signal that can be dropped into existing production systems, routing pipelines, and deployment monitoring infrastructure without modification. Given a model's generated answer, cross-model disagreement computes how surprised or uncertain a second verifier model is when reading that answer via a single forward pass. No generation from the verifying model is required, and no correctness labels are needed. We
instantiate this principle as \gls{cmp}, which measures the verifying
model's surprise at the generating model's answer tokens, and \gls{cme},
which measures the verifying model's uncertainty at those positions.
Both \gls{cmp} and \gls{cme} outperform within-model uncertainty
baselines across benchmarks spanning reasoning, retrieval, and
mathematical problem solving (MMLU, TriviaQA, and GSM8K), and dominate other label-free signals operating in the single-prefill cost regime. On MMLU,
\gls{cmp} achieves a mean AUROC of 0.73 against a within-model
entropy baseline of 0.59. These results establish cross-model disagreement as a practical, training-free approach to label-free correctness estimation, with direct applications in deployment monitoring, model routing, selective prediction,  data filtering, and scalable oversight of production systems.

\end{abstract}

\section{Introduction}

Language models fail in two distinct ways. The first is ignorance: the model does not know the
answer and signals this through uncertainty. Token entropy, maximum softmax probability
\citep{hendrycks2017baseline}, and related signals are reasonable proxies for this failure mode,
and the routing literature has built effective systems around them \citep{ding2024hybrid,
ong2024routellm}. The second failure mode is harder: the model is wrong but certain. It produces
a fluent, high-confidence answer that happens to be incorrect. Within-model uncertainty signals
are blind to this case by construction, since a confident model has low entropy regardless of
whether its answer is right \citep{guo2017calibration}. Yet this is precisely the failure mode
that matters most in practice. A medical assistant confidently stating the wrong drug interaction,
a legal summarizer confidently misreading a statute, a student model confidently propagating a
misconception: these are the errors that cause harm, and existing signals give no warning.

The confident error problem is not merely a calibration issue. Even well-calibrated models
\citep{srinivas2024large} cannot detect their own errors through introspection: a model that is
wrong has, by definition, already committed to a wrong answer. Any signal derived solely from
the generating model's own distribution is fundamentally limited. It can tell you how confident
the model is, but not whether that confidence is warranted. What is needed is an external
perspective: a second model that can evaluate the generating model's answer and flag disagreement.

\begin{figure}[t]
  \centering
  \includegraphics[height=4.7cm]{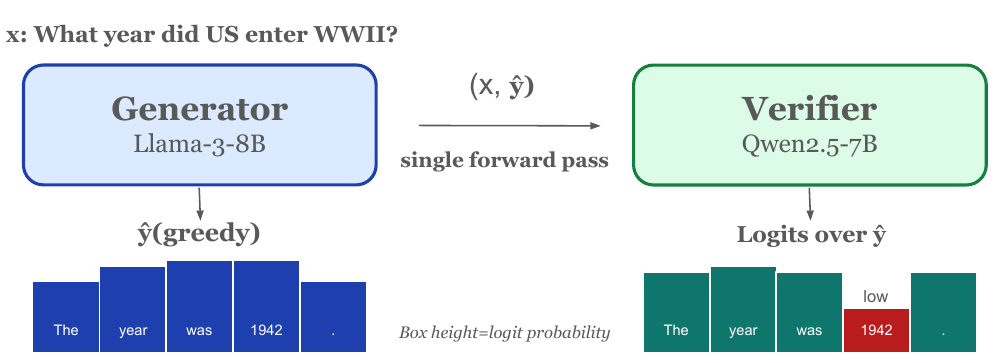}
  \caption{\textbf{Cross-model disagreement as a label-free correctness indicator.}
  Given a prompt $x$, the generator (Llama-3-8B) produces an answer $\hat{y}$. The verifier (Qwen2.5-7B) performs a single forward
  pass over $(x, \hat{y})$ with no generation required. The verifier assigns low
  probability to the token ``1942'', the generator's confident but incorrect
  answer. Cross-model
  perplexity (\gls{cmp}) aggregates this surprise signal into a single
  correctness indicator, and high \gls{cmp} flags a likely error.}
  \label{fig:method}
\end{figure}

This observation motivates \textit{cross-model disagreement} as a practical signal. Rather than asking whether a model is uncertain about its own answer, we ask whether a second model is surprised by it. Concretely, given prompt $x$ and generating model answer $\hat{y}$, we perform a
single forward pass through a verifying model on $(x, \hat{y})$ and extract two
signals: \gls{cmp}, which aggregates the verifying model's token-level surprise,
and \gls{cme}, which aggregates its token-level uncertainty. No generation from the
verifier is required and no correctness labels are needed. The two signals are
complementary: \gls{cmp} is most effective when the generator is confidently wrong
and the verifier assigns low probability to the specific incorrect tokens; \gls{cme}
is more informative on retrieval tasks where distributional uncertainty better
reflects whether an answer is grounded. Across our evaluation, \Gls{cmp} outperforms within-model entropy on MMLU across 12 of 15 model pairs by AUROC, and \gls{cme} leads on TriviaQA where distributional uncertainty is more informative than token-level surprise. When used as a routing signal---directing queries to the verifier only when disagreement is high---\gls{cmp} recovers a substantial fraction of the performance gap between generator and verifier with no labels required, as measured by APGR \citep{ong2024routellm}. \Gls{cmp} wins 14 of 15 routing comparisons against within-model entropy on GSM8K, and achieves mean APGR of 0.803  on MMLU versus 0.546 for G-Ent.

The closest prior work uses cross-model consistency as a hallucination signal.
SelfCheckGPT \citep{manakul2023selfcheckgpt} checks consistency across stochastic samples
from the same model; CrossCheckGPT \citep{sun2024crosscheckgpt} compares outputs generated
by multiple independent models. Both require multiple generations. Our setting is
fundamentally different: we use a single greedy answer and a single forward pass through
the verifier, with no generation from either model after the initial answer, and we target
per-instance correctness prediction rather than hallucination ranking. More broadly, the
uncertainty quantification literature has focused entirely on signals derived from the
generating model itself \citep{kadavath2022language, kuhn2023semantic, liu2020energy}, and
prior work explicitly notes that within-model perplexity fails
\citep{srinivas2024large}. To our knowledge, no prior work uses a verifying model's
logit-based signals on a generating model's answer.

\gls{cmp} and \gls{cme} have direct applications in several settings. In \textit{deployment monitoring}, \gls{cmp} and \gls{cme} can be evaluated on every query without ground truth labels, serving as a cheap triage signal that flags likely errors for more expensive downstream verification---whether human review, re-querying, or generation-based methods such as LLM-as-judge \citep{zheng2023judging}. In \textit{model routing}, high disagreement triggers escalation to a stronger model, recovering a large fraction of the performance gap at a fraction of the cost of always using the strong model. Unlike supervised routers such as RouteLLM \citep{ong2024routellm}, \gls{cmp} and \gls{cme} require no preference labels and no router training, occupying a different point on the supervision-cost tradeoff. In \textit{data filtering}, high \gls{cmp} on a candidate example identifies instances where models disagree---a label-free signal for hard or ambiguous examples useful for training data curation and hard negative mining. Finally, in \textit{selective prediction} \citep{geifman2017selective}, high \gls{cmp} serves as an abstention signal, allowing a system to withhold predictions on inputs where confident errors are most likely — improving accuracy on the subset it does answer without any labeled data or threshold calibration.

We further characterize when cross-model signals are most effective:
on MMLU, \gls{cmp} AUROC is uncorrelated with capability gap
($\rho = 0.11$, $p = 0.72$), suggesting architectural diversity drives
correctness detection rather than capability asymmetry; on TriviaQA,
\gls{cme} is more robust across gap sizes; on GSM8K both signals
improve modestly but without a significant trend. We discuss
implications for scalable oversight and deployment monitoring in
Section~\ref{sec:discussion}.

Our contributions are as follows:
\begin{itemize}
\setlength{\itemsep}{0pt}
\setlength{\parskip}{0pt}
    \item We introduce \textit{cross-model disagreement} as a label-free, training-free correctness
    indicator, instantiated as \gls{cmp} and \gls{cme}, each requiring only a single forward pass
    through a verifying model with no generation or no correctness labels.
    
    \item We show that \gls{cmp} and \gls{cme} outperform within-model uncertainty baselines across MMLU, TriviaQA, and GSM8K, and provide the strongest label-free correctness signal available in the single-prefill compute regime. Methods that achieve higher AUROC (verifier answer agreement, semantic entropy) require autoregressive generation, costing one to two orders of magnitude more per query.

\item We show that on knowledge-intensive tasks, architectural diversity between same-sized
    models is sufficient for effective correctness detection, and capability asymmetry is not
    required, while on open-ended retrieval a stronger verifier provides meaningful additional benefit.

\end{itemize}

\vspace{-1em}
\section{Related Work}
Our work sits at the intersection of LLM uncertainty estimation, cross-model
disagreement, trained verifiers, model routing, scalable oversight, and
speculative decoding.

\textbf{Uncertainty Estimation.}
A standard approach to predicting correctness uses the model's own confidence.
Maximum softmax probability \citep{hendrycks2017baseline}, predictive
entropy, and network features \citep{gorbett2022utilizing} are common baselines, but modern networks are poorly calibrated
\citep{guo2017calibration} and frequently assign high confidence to incorrect
predictions; \citet{srinivas2024large} show explicitly that within-model
perplexity has limited predictive power in open-ended settings. A
complementary line elicits confidence directly, either by prompting for a
yes/no token \citep{kadavath2022language} or a verbalized score
\citep{tian2023just}. Sampling-based methods---self-consistency, semantic
entropy \citep{kuhn2023semantic, farquhar2024detecting}, and ensemble
disagreement \citep{lakshminarayanan2017simple, manakul2023selfcheckgpt,
sun2024crosscheckgpt}---require multiple generation passes. See
\citet{shorinwa2024survey} for a broader taxonomy. \Gls{cmp} and \gls{cme}
require a single verifier forward pass and remain informative precisely
where within-model entropy is flat.

\textbf{Cross-Model Disagreement and Multi-LLM Uncertainty.}
A growing line of work uses signals from a second model to
improve uncertainty estimation beyond what self-consistency provides.
\citet{hamidieh2026complementing} show that cross-model semantic disagreement
among a scale-matched ensemble captures epistemic uncertainty that
self-consistency misses. \citet{xue2025verify} observe that self-consistency
saturates near a black-box oracle and propose a two-stage scheme that invokes
a verifier on uncertain cases. \citet{feng2025rethinking} estimate uncertainty
across multi-agent reformulations of the same query; \citet{dey2025uncertainty}
fuse predictions from multiple LLMs weighted by self-assessment; and
\citet{chen2026modelswitch} switch between LLMs based on cross-model sample
agreement. All require generation from one or more additional models---to
produce response samples, drive multi-agent interaction, or obtain explicit
verifier outputs---and measure agreement between generated answers rather
than logit-level signals on a fixed answer. \Gls{cmp} and \gls{cme} require a
single verifier forward pass with no generation, no sampling, and no
agent-style interaction, producing a scalar signal directly from the
verifier's distribution over the generator's greedy answer.

\textbf{Trained Verifiers and Judges.}
A separate body of work trains a verifier or judge to score candidate
solutions. Outcome and process reward models \citep{
lightman2024lets} assign correctness scores to full solutions or
intermediate steps. Most closely related to our setup, generative verifiers
\citep{zhang2024genrm} reframe verification as next-token prediction: a
verifier trained to assign probability mass to a ``Yes'' or ``No'' token
after the candidate solution. LLM-as-Judge methods
\citep{zheng2023judging, dubois2024alpacafarm, kim2024prometheus} extend
this by having a stronger model generate a scalar rating or structured
critique. \Gls{cmp} differs in two ways: it requires no verifier training,
and the signal is the verifier's existing log-probability over the
generator's actual answer tokens rather than an explicit yes/no judgment or
generated critique. Our P(True) ablation (Appendix~F) confirms that explicit
correctness prompting is far less informative than implicit token-level
perplexity for off-the-shelf verifiers.

\textbf{LLM Routing.}
Routing systems reduce inference cost by directing queries to a stronger
model only when needed. HybridLLM \citep{ding2024hybrid}, RouteLLM
\citep{ong2024routellm}, and FrugalGPT
\citep{chen2023frugalgptuselargelanguage} train routers on labeled
preference data; AutoMix \citep{aggarwal2024automix} uses a smaller model to
self-verify before escalating but still requires a calibration step.
\Gls{cmp} and \gls{cme} are parameter-free and require no router training,
though they require a single forward pass through a second model at
inference time. We adopt the APGR metric from \citet{ong2024routellm} to
measure routing quality.

\textbf{Scalable Oversight.}
Scalable oversight \citep{bowman2022measuring} and weak-to-strong
generalization \citep{burns2023weak} study how to supervise AI systems whose
capabilities may exceed those of the overseer, generally assuming that
effective verification requires a more capable model. We examine whether
capability gap is necessary or whether architectural diversity alone
suffices, evaluating pairs with similar task accuracy alongside asymmetric
ones.

\textbf{Speculative Decoding.}
Speculative decoding \citep{leviathan2023fast, chen2023accelerating}
accelerates inference by using a small draft model to propose tokens, which
a larger target model verifies via acceptance sampling. \Gls{cmp} is the
sequence-level analogue: rather than per-token accept/reject decisions, we
aggregate verifier log-probabilities across the full answer into a single
routing score. This connection grounds why \gls{cmp} is most effective on
tasks where errors manifest as low token-level acceptance probability.
\vspace{-1em}
\section{Method}

In this section we describe cross-model disagreement.
Given a \textit{generating model} $\mathcal{M}_g$ and a \textit{verifying model}
$\mathcal{M}_v$, we define two label-free signals for predicting whether
$\mathcal{M}_g$'s answer is correct.

\textbf{Problem Setup.} Let $x$ denote an input prompt. The generating model produces an answer
$\hat{y}_g \sim \mathcal{M}_g(\cdot \mid x)$ autoregressively by greedy decoding.
Our goal is to predict whether $\hat{y}_g$ is correct without access to ground truth
labels. We make no assumptions about the relative capability of $\mathcal{M}_g$ and
$\mathcal{M}_v$.

\subsection{Cross-Model Perplexity}
Given $\hat{y}_g$, we concatenate the prompt and answer to form $(x, \hat{y}_g)$ and
perform a single forward pass through $\mathcal{M}_v$. At
each answer token position $t \in \{1, \ldots, T\}$, the forward pass yields logits over
the full vocabulary. We define \gls{cmp} as:
\begin{equation}
    \text{CMP}(x, \hat{y}_g) = \exp\left(-\frac{1}{T}\sum_{t=1}^{T}
    \log p_v\!\left(\hat{y}_g^{(t)} \mid x, \hat{y}_g^{(<t)}\right)\right)
\end{equation}
where $p_v$ denotes the verifying model's conditional distribution and $T$ is the number
of answer tokens. High \gls{cmp} indicates that $\mathcal{M}_v$ assigns low probability
to $\mathcal{M}_g$'s answer---the models disagree. \Gls{cmp} connects formally to
speculative decoding \citep{leviathan2023fast, chen2023accelerating}, where the acceptance
probability for a draft token $x_t$ is $\min(1, p_v(x_t) / p_g(x_t))$. \Gls{cmp}
aggregates this token-level acceptance signal across the full answer sequence, making a
single binary decision rather than a token-level correction.

\subsection{Cross-Model Entropy}

From the same single forward pass, we also compute \gls{cme} as the mean entropy of the
verifying model's output distribution over answer token positions:

\begin{equation}
    \text{CME}(x, \hat{y}_g) = -\frac{1}{T}\sum_{t=1}^{T}\sum_{v}
    p_v(v \mid x, \hat{y}_g^{(<t)}) \log p_v(v \mid x, \hat{y}_g^{(<t)})
\end{equation}

Where \gls{cmp} measures the verifying model's surprise at the specific tokens produced,
\gls{cme} measures its general uncertainty at those positions. The two signals are
complementary: \gls{cmp} is most informative when the generator is confidently wrong and
the verifier assigns low probability to the specific incorrect tokens; \gls{cme} is most
informative on retrieval tasks where the verifier's distributional uncertainty better
reflects whether the answer is grounded. Both signals are obtained at no additional
computational cost beyond the single forward pass required for \gls{cmp}.

\begin{figure}[t]
    \centering
    \includegraphics[width=\textwidth]{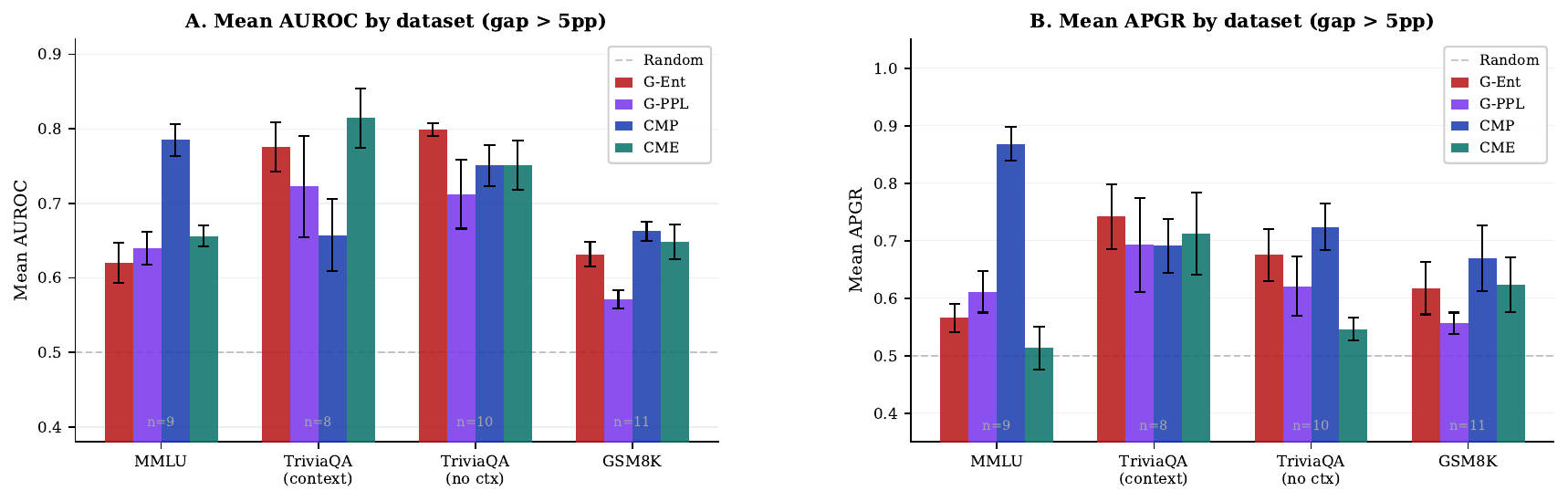}
    \caption{\textbf{\gls{cmp} and \gls{cme} performance across datasets compared to baselines.} \textbf{(A)}~Mean AUROC over all model pairs. G-Ent and G-PPL measure the generator's own entropy and perplexity; CME and CMP measure the corresponding signals from a verifier model on the generator's answer.  APGR measures the fraction of the performance gap between weak and strong model that is recovered by routing, normalized so that random routing scores 0 and oracle routing scores 1; pairs with small gaps are excluded because the small denominator in the PGR formula produces unstable estimates. Error bars show standard error across pairs; $n$ indicates the number of pairs per dataset. }
    \label{fig:bar-apgr}
\end{figure}

\vspace{-1em}
\section{Experiments}
\vspace{-.5em}

We evaluate \gls{cmp} and \gls{cme} across four benchmarks datasets. Table~\ref{tab:main} summarizes selected
results; full results appear in Appendix~\ref{app:full_results}.

\textbf{Datasets.}
We evaluate on MMLU \citep{hendrycks2021measuring} (multiple-choice),
TriviaQA \citep{joshi2017triviaqa} (with and without context), and
GSM8K \citep{cobbe2021gsm8k} (chain-of-thought reasoning), covering
knowledge retrieval, reading comprehension, and multi-step arithmetic.

\textbf{Models.}
We evaluate seven instruction-tuned models spanning five families: Qwen2.5 (0.5B, 7B), Llama-3 (1B, 8B), Gemma-3 (270M), Mistral (7B), and OLMo-3 (7B). We construct ordered model pairs across four datasets, covering both \textit{asymmetric} pairs (small generator verified by large model, e.g.\ Qwen-0.5B $\to$ Qwen-7B) and \textit{same-sized cross-family} pairs among the four 7--8B models (Qwen-7B, Llama-3-8B, Mistral-7B, OLMo-7B), which isolate architectural diversity from capability asymmetry. Full model details and per-pair results are in Appendix~\ref{app:setup}.

\textbf{Within-model baselines and metrics.}
We compare \gls{cmp} and \gls{cme} against two within-model baselines:
\textit{generator entropy} (G-Ent), the mean token-level entropy
$H = -\sum_v p_v \log p_v$ over the generator's answer tokens, and
\textit{generator perplexity} (G-PPL), the mean token-level perplexity
of the generator on its own answer. Both are standard unsupervised signals
requiring no verifier. Together these serve as direct
ablations of \gls{cme} and \gls{cmp} respectively: if the cross-model
signals merely recover information already present in the generator's
own distribution, G-Ent and G-PPL should perform comparably. We additionally test explicit verification prompting
(P(True); \citealp{kadavath2022language}) on GSM8K, where the verifier
is asked directly whether the generator's answer is correct
(Appendix~\ref{sec:gsm8k-ablation}). We also compare against
RouteLLM \citep{ong2024routellm}, the strongest supervised routing
baseline, which trains a router on human preference labels augmented
with LLM-judge data; unlike our signals, it requires labeled data and
a trained router model, and we treat it as an upper bound on what
supervision buys in this setting.

For correctness prediction we report AUROC against weak model
incorrectness, which is threshold-free, requires no routing setup, and
is our primary metric for evaluating signal quality across all pairs.
For routing quality we report APGR (Average Performance Gap Recovered)
\citep{ong2024routellm}. At each routing threshold, a fraction $c$ of
queries are sent to the strong model while the rest use the weak
model's answer, yielding accuracy $a(c)$. The performance gap recovered
at cost $c$ is:
\begin{equation}
    \text{PGR}(c) = \frac{a(c) - a_w}{a_s - a_w},
\end{equation}
where $a_w$ and $a_s$ are the weak and strong model accuracies. APGR
averages PGR over $c \in (0, 1)$: a value of 0 corresponds to random
routing and 1 to oracle routing. Small accuracy gaps produce unstable
APGR estimates due to the small denominator in $\text{PGR}(c)$; we
therefore exclude such pairs from Figure~\ref{fig:bar-apgr}.

\begin{table*}[t]
\centering
\scriptsize
\setlength{\tabcolsep}{3.5pt}
\caption{Selected results across all four benchmarks. \textbf{Bold} = best per row separately for AUROC and APGR among G-Ent, G-PPL, CMP, CME. Same-size cross-family pairs marked $\dagger$. See Appendix for full results.}
\label{tab:main}
\begin{tabular}{llcccc | cccc | cccc}
\toprule
& & & & & Acc. & \multicolumn{4}{c|}{AUROC $\uparrow$} & \multicolumn{4}{c}{APGR $\uparrow$} \\
Dataset & Generator & Verifier & Acc$_g$ & Acc$_v$ & Gap & G-Ent & G-PPL & CMP & CME & G-Ent & G-PPL & CMP & CME \\
\midrule
\multirow{6}{*}{MMLU} & Qwen-0.5B & Qwen-7B & 0.42 & 0.72 & 0.30 & 0.588 & 0.556 & \textbf{0.841} & 0.583 & 0.521 & 0.501 & \textbf{0.793} & 0.497 \\
 & Llama-1B & Llama-3-8B & 0.43 & 0.62 & 0.19 & 0.654 & 0.655 & \textbf{0.817} & 0.679 & 0.562 & 0.605 & \textbf{0.869} & 0.540 \\
 & Llama-1B & Qwen-7B & 0.43 & 0.72 & 0.29 & 0.654 & 0.655 & \textbf{0.786} & 0.634 & 0.569 & 0.595 & \textbf{0.807} & 0.558 \\
 & OLMo-7B$^\dagger$ & Qwen-7B & 0.56 & 0.72 & 0.15 & 0.503 & 0.599 & \textbf{0.896} & 0.697 & 0.486 & 0.546 & \textbf{0.914} & 0.447 \\

\midrule
\multirow{5}{*}{TriviaQA} & Llama-3-8B & Mistral-7B & 0.43 & 0.64 & 0.20 & 0.859 & 0.885 & 0.825 & \textbf{0.922} & 0.822 & 0.831 & \textbf{0.849} & 0.789 \\
 & Qwen-0.5B & Qwen-7B & 0.41 & 0.65 & 0.24 & 0.778 & 0.730 & 0.816 & \textbf{0.825} & 0.621 & 0.604 & \textbf{0.738} & 0.610 \\
 & Llama-1B & Qwen-7B & 0.55 & 0.65 & 0.10 & 0.804 & \textbf{0.810} & 0.688 & 0.789 & 0.691 & 0.720 & \textbf{0.761} & 0.574 \\
 & Llama-3-8B$^\dagger$ & Qwen-7B & 0.43 & 0.65 & 0.21 & 0.859 & \textbf{0.885} & 0.772 & 0.875 & 0.796 & \textbf{0.808} & 0.807 & 0.757 \\
\midrule
\multirow{4}{*}{\shortstack{TriviaQA\\{\scriptsize(no ctx)}}} & Llama-1B & Llama-3-8B & 0.42 & 0.67 & 0.25 & 0.840 & 0.851 & \textbf{0.938} & 0.814 & 0.685 & 0.693 & \textbf{0.811} & 0.589 \\
 & Qwen-0.5B & Qwen-7B & 0.21 & 0.59 & 0.38 & 0.782 & 0.775 & 0.793 & \textbf{0.817} & 0.530 & 0.532 & \textbf{0.644} & 0.510 \\
 & Llama-1B & Qwen-7B & 0.42 & 0.59 & 0.17 & 0.840 & \textbf{0.851} & 0.738 & 0.806 & 0.653 & 0.671 & \textbf{0.759} & 0.501 \\
 & Qwen-7B$^\dagger$ & Llama-3-8B & 0.59 & 0.67 & 0.08 & \textbf{0.823} & 0.793 & 0.748 & 0.625 & \textbf{0.980} & 0.946 & 0.964 & 0.533 \\
\midrule
\multirow{6}{*}{GSM8K} & Mistral-7B$^\dagger$ & Llama-3-8B & 0.54 & 0.61 & 0.07 & 0.658 & 0.542 & \textbf{0.704} & 0.675 & 0.955 & 0.663 & \textbf{1.155} & 0.977 \\
 & Llama-1B & Qwen-7B & 0.37 & 0.90 & 0.53 & 0.559 & 0.531 & 0.652 & \textbf{0.659} & 0.515 & 0.506 & \textbf{0.563} & 0.556 \\
 & Llama-1B & Llama-3-8B & 0.37 & 0.61 & 0.25 & 0.559 & 0.531 & \textbf{0.653} & 0.614 & 0.536 & 0.517 & \textbf{0.633} & 0.589 \\
 & Qwen-0.5B & Qwen-7B & 0.24 & 0.90 & 0.66 & 0.640 & 0.630 & 0.664 & \textbf{0.698} & 0.527 & 0.528 & 0.537 & \textbf{0.543} \\

\bottomrule
\end{tabular}
\end{table*}
\vspace{-1em}
\subsection{Results}\label{sec:results}

We evaluate \gls{cmp} and \gls{cme} against within-model baselines across
four benchmarks and model pairs spanning asymmetric and same-sized
cross-family configurations. Table~\ref{tab:main} and
Figure~\ref{fig:bar-apgr} summarize the main results: \gls{cmp} leads
on MMLU and GSM8K while \gls{cme} is more competitive on TriviaQA,
and both cross-model signals consistently outperform their within-model
counterparts on tasks where the generator makes confident errors. The
paragraphs below analyze the per-case signal structure, quintile
separation, and conditions under which cross-model signals succeed or
fail.

\begin{figure*}[t]
\centering
\includegraphics[width=\linewidth]{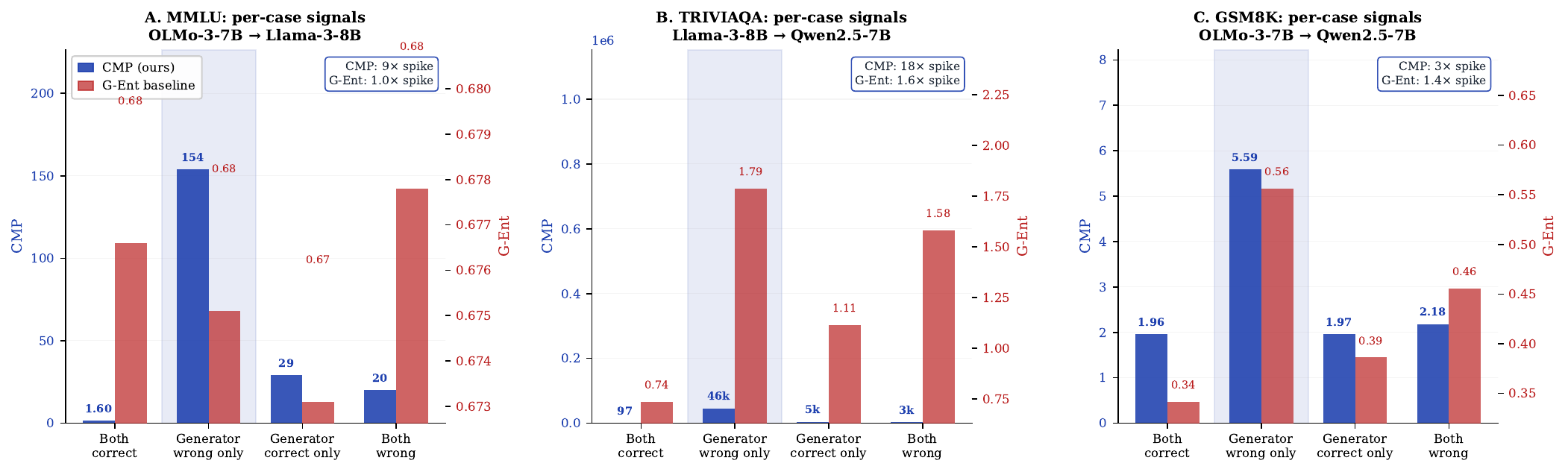}\\[0.8em]
\includegraphics[width=\linewidth]{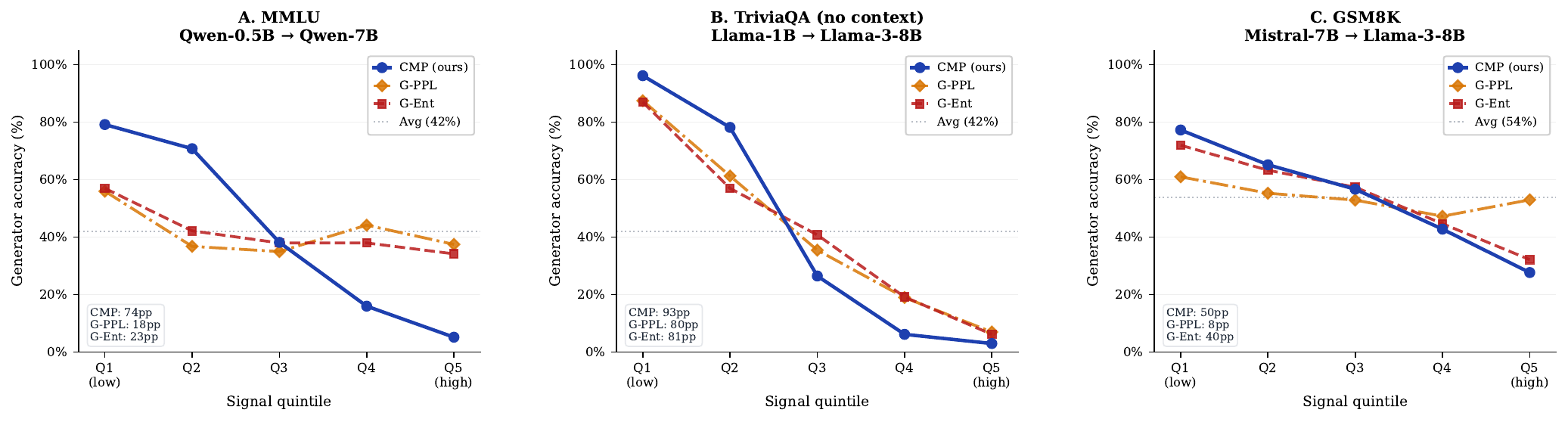}
\caption{\textbf{Top row: per-case signal means.} Mean \gls{cmp} (left
axis, blue) and G-Ent (right axis, red) by outcome category. The shaded
column highlights the ``generator wrong only'' case---confident errors
the verifier does not share. On MMLU, \gls{cmp} spikes $9\times$ above
the mean of the other three cases while G-Ent is flat ($1.0\times$);
on TriviaQA the spike is $18\times$ vs.\ $1.6\times$; on GSM8K both
signals rise modestly ($3\times$ and $1.4\times$), reflecting the
difficulty of isolating chain-of-thought errors with token-level signals.
\textbf{Bottom row: accuracy by signal quintile.} Samples sorted by
signal strength (Q1 = lowest, Q5 = highest); bars show weak model
accuracy within each bin. On MMLU, \gls{cmp} produces a 74pp spread
versus 23pp for G-Ent and 18pp for G-PPL. On TriviaQA (no context),
all three signals are competitive (93pp, 80pp, 81pp). On GSM8K,
\gls{cmp} achieves a 50pp spread while G-PPL nearly collapses to 8pp,
confirming that generator self-perplexity is uninformative on
chain-of-thought tasks and that cross-model disagreement is doing
genuine work.}
\label{fig:analysis}
\end{figure*}

\paragraph{CMP targets the failure mode G-Ent misses.}
Figure~\ref{fig:analysis} isolates the four outcome categories.
On MMLU (OLMo-7B $\to$ Llama-8B), \gls{cmp} in the ``generator wrong
only'' case is $154\times$ the both-correct baseline---a 9$\times$ spike
over the remaining cases---while G-Ent is flat across all four ($\leq
1.0\times$ variation). On TriviaQA (Llama-8B $\to$ Qwen-7B) the pattern
sharpens: \gls{cmp} reaches $46\mathrm{k}$ versus $97$ when both models
are correct (18$\times$ selectivity), while G-Ent rises by only
1.6$\times$. G-Ent is elevated whenever the generator is uncertain
regardless of whether the verifier concurs; \gls{cmp} selectively spikes
only when the generator's error is not shared by the verifier.

Figure~\ref{fig:bar-apgr}A reports mean AUROC across all model pairs.
On MMLU, \gls{cmp} wins 12 of 15 pairs by AUROC, with a mean of 0.727
versus 0.595 for G-Ent and 0.607 for G-PPL. On GSM8K, \gls{cmp} (0.623)
and \gls{cme} (0.621) both outperform G-Ent (0.584) and G-PPL (0.533),
with G-PPL performing worst of all four signals---consistent with the
near-flat quintile spread in Figure~\ref{fig:analysis}. On TriviaQA
(context), \gls{cme} leads (0.743) with G-Ent close behind (0.732),
while \gls{cmp} underperforms on several pairs where the verifier
shares similar knowledge gaps to the generator. On TriviaQA (no
context) the AUROC signals cluster (G-Ent~0.779, CMP~0.750, CME~0.738,
G-PPL~0.687): G-Ent is the strongest AUROC signal on this dataset, while
\gls{cmp} leads on routing-relevant APGR, consistent with retrieval
errors being detectable by within-model signals when context is absent.

\paragraph{Task-dependent boundaries.}
On GSM8K, \gls{cmp} achieves mean APGR of 0.628 versus 0.583 for G-Ent, winning 14 of 15 routing comparisons against G-Ent, but the per-case spike
is weaker than on MMLU or TriviaQA, and \gls{cme} is preferable on
several pairs (Table~\ref{tab:main}). The Mistral-7B $\to$ Qwen-7B pair
on TriviaQA is a more extreme case: with only a 1pp accuracy gap,
\gls{cmp} and G-PPL collapse near chance (AUROC 0.42 and 0.30) while
G-Ent remains strong (0.847). Both cases point to the same boundary
condition: when the verifier shares the generator's failure
mode---either through similar knowledge gaps or nearly identical
accuracy---cross-model disagreement loses its signal, and within-model
entropy is the more reliable fallback.

\paragraph{Final-answer tokens vs.\ full chain-of-thought.}
On GSM8K, we test whether \gls{cmp}'s signal comes from the full chain-of-thought or only the final numerical answer. \gls{cmp} restricted to answer tokens only (CMP-Final) achieves mean APGR of 0.665 versus 0.682 for the full trace (CMP-Full), indicating that the final answer carries most of the signal with a small additional benefit from the reasoning steps. This suggests \gls{cmp} could be applied more efficiently by scoring only the answer tokens, avoiding the cost of processing the full chain-of-thought through the verifier.


\paragraph{Comparison to supervised routing.} RouteLLM \citep{ong2024routellm} provides the nearest supervised baseline: a causal LLM classifier trained on tens of thousands of human preference labels from Chatbot Arena, augmented with LLM-judge synthetic labels, routing between GPT-4 and Mixtral-8x7B. Their best augmented router achieves APGR of 0.622 on GSM8K and 0.603 on MMLU. On the 8-pair GSM8K subset used for our chain-of-thought ablation (Appendix~\ref{sec:gsm8k-ablation}), CMP-Full achieves APGR of 0.682; over the full set of pairs with measurable capability gap ($\text{gap} > 0.05$), \gls{cmp} achieves mean APGR of 0.628 on GSM8K and 0.803 on MMLU with zero labeled data and no router training. These evaluations are not directly comparable: RouteLLM routes between a proprietary strong model and an open-weight weak model across a different task and label distribution, while we route between open-weight pairs on standard benchmarks. We include the comparison not as a controlled benchmark but to situate \gls{cmp} on the broader supervision-cost tradeoff: a single verifier prefill with no labeled data is competitive with a trained supervised router on reasoning tasks, and the gap is substantial on knowledge-intensive multiple-choice. Where labeled routing data can be collected, supervised routers remain the appropriate choice.

\begin{table}[t]
\centering
\small
\caption{Mean AUROC across model pairs with capability gap $>$ 5pp,
comparing CMP and CME against label-free baselines. Methods are grouped
by computational cost tier; cost is per-sample overhead relative to a
single verifier prefill, excluding shared generator inference.
$T$ = answer tokens; $k{=}10$ for Semantic Entropy.
\textbf{Bold} = best per column within tier; \underline{underline} = best
overall per column. Full per-pair results in Appendix~\ref{tab:combined_auroc}.}
\label{tab:baseline-summary}
\begin{tabular}{lccccll}
\toprule
 & \multicolumn{4}{c}{Mean AUROC $\uparrow$} & & \\
\cmidrule(lr){2-5}
Method & MMLU & TriviaQA & TriviaQA & GSM8K & Verifier work & Cost \\
       &      & (ctx)    & (no ctx) &       &               & vs.\ CMP \\
\midrule
\multicolumn{7}{l}{\emph{Single verifier prefill}} \\
P(True)            & 0.518          & 0.510          & 0.526          & 0.603          & 1 prefill         & $\sim$1$\times$ \\
CME (ours)         & 0.628          & \textbf{0.759} & 0.737          & 0.621          & 1 prefill         & 1$\times$ \\
CMP (ours)         & \textbf{0.749} & 0.679          & \textbf{0.761} & \textbf{0.623} & 1 prefill         & 1$\times$ \\
\midrule
\multicolumn{7}{l}{\emph{Autoregressive generation required}} \\
V-Agree            & \underline{0.756} & \underline{0.782} & 0.789             & \underline{0.898} & 1 gen ($T$ tok)            & 1--$T\times$ \\
Sem-Ent ($k{=}10$) & 0.611             & 0.747             & \underline{0.804} & 0.728             & $k$ gens + NLI             & $k$--$kT\times$ \\
\bottomrule
\end{tabular}
\end{table}

\subsection{Comparison to Baselines}
\label{sec:baselines}
To situate CMP and CME against a broader range of label-free correctness
signals, we evaluate three baselines that vary the source and
form of disagreement. Each baseline targets a distinct point on the
supervision-cost spectrum and isolates a different aspect of what the
verifier contributes.

\textbf{Verifier Answer Agreement} (V-Agree) uses string-level
disagreement between greedy answers from $M_g$ and $M_v$, ablating
verifier logits while preserving cross-model structure. \textbf{P(True)}
\citep{kadavath2022language} prompts the verifier with the generator's answer
and extracts the probability on the ``Yes'' token, replacing CMP's
implicit token-level perplexity with an explicit verification judgment
in the same single-prefill regime. \textbf{Semantic Entropy}
\citep{kuhn2023semantic} removes the verifier entirely, drawing $k$ generator
samples at $T > 0$, clustering them via an NLI model, and computing
entropy over clusters; it is the strongest within-model consistency
baseline in the uncertainty quantification literature. The three
baselines span a range of compute regimes from a single verifier prefill
(P(True)) to $k$ full generations plus an auxiliary NLI model
(Semantic Entropy); Table~\ref{tab:baseline-summary} summarizes
per-sample cost alongside results.

\textbf{Results} Among single-prefill signals, CMP or CME achieves the highest AUROC on every benchmark, with the choice between them tracking task structure: CMP wins on tasks where errors are token-localized (MMLU multiple-choice, TriviaQA without context), CME wins on retrieval. P(True), despite operating in the same compute regime, performs near chance — consistent with known limitations of explicit self-evaluation prompting for off-the-shelf verifiers. Generation-based baselines (V-Agree, Sem-Ent) achieve higher AUROC on TriviaQA (no context) and GSM8K, but at substantially higher cost: V-Agree requires autoregressive decode steps from the verifier, and Sem-Ent requires $k$ full generations from the generator plus an auxiliary NLI model. CMP and CME occupy the cheapest point on the cost-quality curve.

\vspace{-1em}

\section{Discussion}\label{sec:discussion}


\subsection{Routing performance and compute tradeoffs}
Figure~\ref{fig:bar-apgr}B reports mean APGR across pairs with accuracy
gap $>$ 5pp. \Gls{cmp} leads on MMLU and GSM8K; on TriviaQA, G-Ent
and \gls{cme} are competitive, consistent with the AUROC results.

The practical regime where \gls{cmp} is most attractive is one where
generation cost dominates and no labeled routing data is available.
Routing with \gls{cmp} or \gls{cme} requires a single verifier prefill
on every query, with no autoregressive generation from the verifier and
no router model. On MMLU this prefill is negligible; on GSM8K, where
chain-of-thought traces reach up to 256 tokens, it is more substantial
but still well below the cost of strong-model generation. Critically,
the prefill is incurred only to make a routing decision: queries
correctly identified as easy are served by the weak model with no
further large-model compute. If the verifier is generated rather than
prefilled, verifier compute would be incurred on every query regardless
of the routing outcome, largely eliminating the cost benefit.
Prefill-only is therefore what makes label-free routing economically
viable on long-form tasks, not a concession.

Existing approaches occupy different points on this tradeoff.
Label-based routers such as RouteLLM \citep{ong2024routellm} and
HybridLLM \citep{ding2024hybrid} add negligible inference overhead but
require labeled preference data and a trained router model. FrugalGPT
\citep{chen2023frugalgptuselargelanguage} and AutoMix
\citep{aggarwal2024automix} avoid large-model compute on easy queries
entirely but require calibration steps or supervised signals. \Gls{cmp}
and \gls{cme} occupy the opposite corner: zero labeled data, zero router
training, one verifier prefill per query. The tradeoff is explicit:
where labeled routing data can be collected, supervised routers remain
the appropriate choice; \gls{cmp} and \gls{cme} are most valuable
precisely when such data is unavailable.

\subsection{When Does Capability Gap Matter?}

\begin{figure*}[t]
\centering
\includegraphics[width=\linewidth]{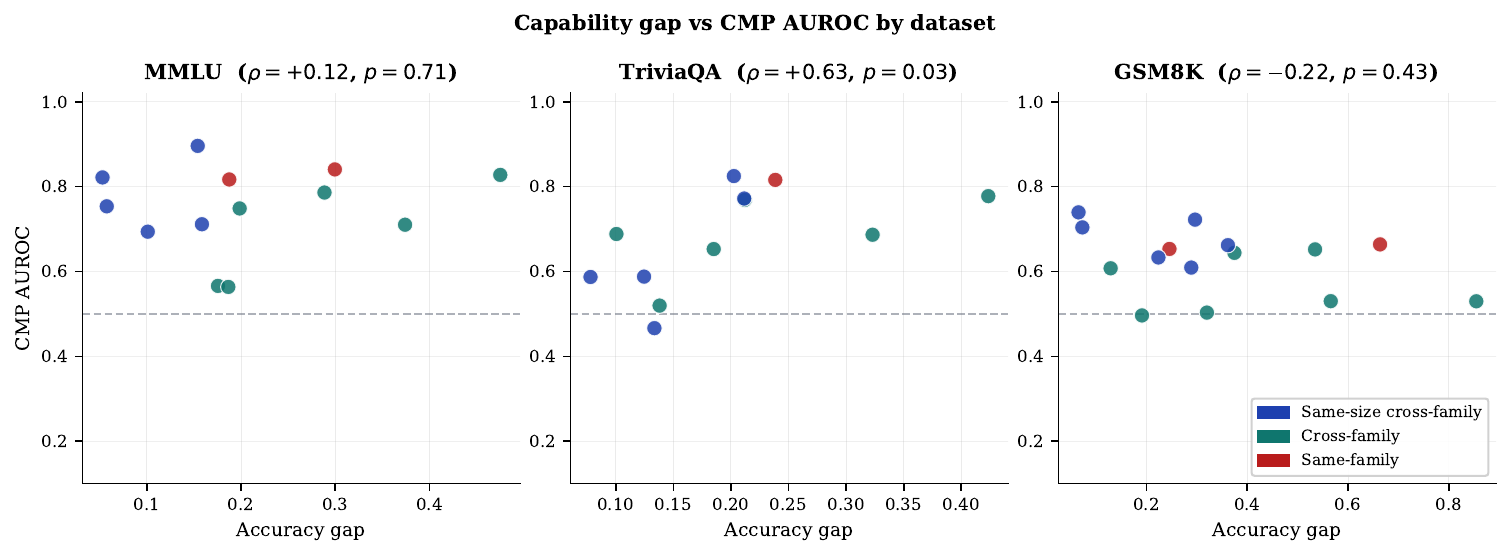}
\caption{\Gls{cmp} AUROC versus capability gap across three benchmarks.
On MMLU, AUROC is uncorrelated with gap ($\rho = +0.12$, $p = 0.71$)
and same-size cross-family pairs (blue) achieve the highest scores,
suggesting model diversity drives the signal rather than capability
asymmetry. On TriviaQA, gap correlates positively with AUROC
($\rho = +0.63$, $p = 0.03$), indicating a stronger verifier helps
when errors are knowledge-driven. GSM8K shows no significant trend
($\rho = -0.22$, $p = 0.43$).}
\label{fig:gap_vs_auroc}
\end{figure*}

Figure~\ref{fig:gap_vs_auroc} plots \gls{cmp} AUROC against accuracy
gap for all pairs across three benchmarks. The task-dependent pattern
is stark. On MMLU, \gls{cmp} is essentially uncorrelated with
capability gap ($\rho = +0.12$, $p = 0.71$): pairs with a 6-point
accuracy gap perform as well as pairs with a 47-point gap, and the
highest-scoring points are same-size cross-family pairs (blue),
not the largest-gap pairs. The operative mechanism on
knowledge-intensive multiple-choice tasks appears to be diversity of
training rather than capability asymmetry---models from different
families make different confident errors, and \gls{cmp} captures this
disagreement regardless of relative accuracy.

On TriviaQA the picture is different. \Gls{cmp} shows a significant
positive correlation with capability gap ($\rho = +0.63$, $p = 0.03$),
with large-gap pairs substantially outperforming small-gap pairs
(AUROC 0.78 vs.\ 0.59). Open-ended knowledge retrieval requires the
verifier to have genuine knowledge the generator lacks; a same-sized
peer from a different family may share the same knowledge gaps,
reducing the informativeness of its surprise. The two same-size
cross-family pairs on TriviaQA (blue) sit at the lower end of the
performance range, consistent with this interpretation. \Gls{cme} is
more robust to this effect, remaining competitive across gap sizes.

On GSM8K there is no significant trend ($\rho = -0.22$, $p = 0.43$),
with high variance across pairs at similar gap sizes. Signal quality
on chain-of-thought tasks appears driven more by model family and
architectural diversity than by capability gap, though the modest
overall AUROC levels make these patterns harder to interpret cleanly.

\subsection{Implications for Scalable Oversight}

The scalable oversight literature \citep{bowman2022measuring,
burns2023weak} has largely assumed that verifying a model's outputs
requires a more capable supervisor. Our results qualify this assumption
in a task-dependent way. On MMLU, peer verification between same-sized
models of different families is as effective as verification by a
substantially stronger model, suggesting that for knowledge-intensive
tasks the capability hierarchy is not a prerequisite for correctness
detection --- architectural diversity is sufficient. On open-ended
retrieval tasks, a stronger verifier does provide meaningful benefit,
consistent with the intuition that the verifier needs knowledge the
generator lacks.
 
The practical implication is straightforward: the right verification
strategy depends on the likely failure mode of the deployed model. For
knowledge-intensive tasks with constrained output formats, a diverse
peer model is a sufficient and efficient verifier. For open-ended
retrieval, a stronger verifier improves detection quality. This
task-dependent characterization is a concrete and actionable finding
for deployment monitoring without labeled data.

\vspace{-1em}

\section{Conclusion}
We have shown that cross-model disagreement---measured as \gls{cmp} or
\gls{cme} via a single forward pass through a verifying model---is a
reliable label-free correctness signal for language model outputs.
The signal works between same-sized models of different families and
requires no labeled data or router training.

\paragraph{Limitations.}
CMP and CME require the verifier's distribution to be a better proxy
for correctness than the generator's. All evaluated pairs have
non-negative capability gap; the reverse case (weaker verifier) is
not directly tested, though our boundary analysis on Mistral$\to$Qwen
TriviaQA (1pp gap, CMP AUROC 0.42) suggests signal degrades as the
gap shrinks or reverses. Token-level scoring is also sensitive to
paraphrase: when the verifier knows a wording the generator did not
produce, CMP can penalize the phrasing rather than the answer, and
CME is more robust to this.

\paragraph{Future work.}
Several directions remain open. Extending cross-model disagreement
beyond greedy decoding to sampled or long-form chain-of-thought
outputs --- and to code generation and multi-step reasoning at scale
--- would test the framework's limits. A theoretical account of which
signal (CMP vs.\ CME) works for which failure mode would deepen the
empirical characterization we provide. The diversity mechanism we
identify raises a concrete design question: given a fixed verifier
budget, how should one select a verifying model to maximize
correctness signal? Representational similarity metrics may offer a
principled answer. Finally, our scalable oversight results are
inference-time on static benchmarks; whether peer verification
supports iterative oversight loops as both models scale is the
central open question connecting this work to the broader alignment
agenda.

\bibliography{colm2026_conference}
\bibliographystyle{colm2026_conference}

\clearpage
\appendix
\section*{Appendix}
\section{Full Results} \label{app:full_results}

Tables~\ref{tab:mmlu}--\ref{tab:gsm8k} report per-pair results across all four
benchmarks. Pairs are grouped by type---same-size cross-family, cross-family
asymmetric, and same-family asymmetric---and sorted by capability gap (descending)
within each group. For each pair we report AUROC and APGR for the within-model
entropy baseline (G-Ent), within-model perplexity (G-PPL), cross-model perplexity (CMP), and cross-model entropy
(CME). \textbf{Bold} indicates the best signal per row (separately for the AUROC
and APGR groups). Because APGR has the form $(a(c) - a_w)/(a_s - a_w)$ and its
denominator becomes unstable when the generator--verifier accuracy gap is small,
we report two APGR mean rows where applicable: one over all pairs, and one
restricted to pairs with accuracy gap $> 0.05$. Rows with $\text{gap} \le 0.05$ are marked $^{\dagger}$.

\begin{table*}[h]
\centering
\scriptsize
\setlength{\tabcolsep}{3pt}
\caption{Full results on MMLU. Pairs grouped by type and sorted by capability gap (descending). $^{\dagger}$ marks rows with $\text{gap} \le 0.05$, which produce unstable APGR estimates due to the small PGR denominator and are excluded from the filtered mean. \textbf{Bold} = best per row among G-Ent, G-PPL, CMP, CME.}
\label{tab:mmlu}
\begin{tabular}{llccc | cccc | cccc}
\toprule
& & & & & \multicolumn{4}{c|}{AUROC $\uparrow$} & \multicolumn{4}{c}{APGR $\uparrow$} \\
Generator & Verifier & Acc$_g$ & Acc$_v$ & Gap & G-Ent & G-PPL & CMP & CME & G-Ent & G-PPL & CMP & CME \\
\midrule
\multicolumn{13}{l}{\textit{Same-size cross-family}} \\
Mistral-7B & Qwen-7B & 0.56 & 0.72 & 0.16 & 0.720 & \textbf{0.726} & 0.711 & 0.632 & 0.608 & 0.662 & \textbf{0.818} & 0.605 \\
OLMo-7B & Qwen-7B & 0.56 & 0.72 & 0.15 & 0.503 & 0.599 & \textbf{0.896} & 0.697 & 0.486 & 0.546 & \textbf{0.914} & 0.447 \\
Llama-3-8B & Qwen-7B & 0.62 & 0.72 & 0.10 & 0.652 & 0.684 & \textbf{0.694} & 0.635 & 0.617 & 0.652 & \textbf{0.910} & 0.603 \\
Mistral-7B & Llama-3-8B & 0.56 & 0.62 & 0.06 & 0.720 & 0.726 & \textbf{0.753} & 0.716 & 0.715 & 0.853 & \textbf{0.941} & 0.629 \\
OLMo-7B & Llama-3-8B & 0.56 & 0.62 & 0.05 & 0.503 & 0.599 & \textbf{0.822} & 0.689 & 0.483 & 0.582 & \textbf{1.026} & 0.259 \\
Mistral-7B & OLMo-7B$^{\dagger}$ & 0.56 & 0.56 & 0.00 & 0.720 & \textbf{0.726} & 0.584 & 0.605 & 3.926 & \textbf{5.296} & 1.358 & 3.778 \\
\midrule
\multicolumn{13}{l}{\textit{Cross-family asymmetric}} \\
Gemma-270M & Qwen-7B & 0.24 & 0.72 & 0.47 & 0.509 & 0.515 & \textbf{0.828} & 0.609 & 0.502 & 0.504 & \textbf{0.673} & 0.534 \\
Gemma-270M & Llama-3-8B & 0.24 & 0.62 & 0.37 & 0.509 & 0.515 & \textbf{0.710} & 0.597 & 0.506 & 0.501 & \textbf{0.653} & 0.524 \\
Llama-1B & Qwen-7B & 0.43 & 0.72 & 0.29 & 0.654 & 0.655 & \textbf{0.786} & 0.634 & 0.569 & 0.595 & \textbf{0.807} & 0.558 \\
Qwen-0.5B & Llama-3-8B & 0.42 & 0.62 & 0.20 & 0.588 & 0.556 & \textbf{0.749} & 0.639 & 0.532 & 0.503 & \textbf{0.740} & 0.482 \\
Gemma-270M & Llama-1B & 0.24 & 0.43 & 0.19 & 0.509 & 0.515 & \textbf{0.564} & 0.532 & 0.497 & 0.516 & \textbf{0.616} & 0.417 \\
Gemma-270M & Qwen-0.5B & 0.24 & 0.42 & 0.18 & 0.509 & 0.515 & \textbf{0.566} & 0.526 & 0.506 & 0.486 & \textbf{0.673} & 0.421 \\
Qwen-0.5B & Llama-1B$^{\dagger}$ & 0.42 & 0.43 & 0.01 & 0.588 & 0.556 & 0.580 & \textbf{0.605} & 1.409 & 1.303 & \textbf{1.773} & 0.394 \\
\midrule
\multicolumn{13}{l}{\textit{Same-family asymmetric}} \\
Qwen-0.5B & Qwen-7B & 0.42 & 0.72 & 0.30 & 0.588 & 0.556 & \textbf{0.841} & 0.583 & 0.521 & 0.501 & \textbf{0.793} & 0.497 \\
Llama-1B & Llama-3-8B & 0.43 & 0.62 & 0.19 & 0.654 & 0.655 & \textbf{0.817} & 0.679 & 0.562 & 0.605 & \textbf{0.869} & 0.540 \\
\midrule
\textit{Mean (all, $n{=}15$)} &  &  &  &  & 0.595 & 0.607 & \textbf{0.727} & 0.625 & 0.829 & \textbf{0.940} & 0.904 & 0.712 \\
\textit{Mean (gap${>}0.05$, $n{=}13$)} &  &  &  &  & 0.586 & 0.601 & \textbf{0.749} & 0.628 & 0.546 & 0.577 & \textbf{0.803} & 0.501 \\
\bottomrule
\end{tabular}
\end{table*}

\vspace{2em}

\begin{table*}[h]
\centering
\scriptsize
\setlength{\tabcolsep}{3pt}
\caption{Full results on TriviaQA (with context). Pairs grouped by type and sorted by capability gap (descending). $^{\dagger}$ marks rows with $\text{gap} \le 0.05$. \textbf{Bold} = best per row among G-Ent, G-PPL, CMP, CME.}
\label{tab:triviaqa}
\begin{tabular}{llccc | cccc | cccc}
\toprule
& & & & & \multicolumn{4}{c|}{AUROC $\uparrow$} & \multicolumn{4}{c}{APGR $\uparrow$} \\
Generator & Verifier & Acc$_g$ & Acc$_v$ & Gap & G-Ent & G-PPL & CMP & CME & G-Ent & G-PPL & CMP & CME \\
\midrule
\multicolumn{13}{l}{\textit{Same-size cross-family}} \\
Llama-3-8B & Qwen-7B & 0.43 & 0.65 & 0.21 & 0.859 & \textbf{0.885} & 0.772 & 0.875 & 0.796 & \textbf{0.808} & 0.807 & 0.757 \\
Llama-3-8B & Mistral-7B & 0.43 & 0.64 & 0.20 & 0.859 & 0.885 & 0.825 & \textbf{0.922} & 0.822 & 0.831 & \textbf{0.849} & 0.789 \\
OLMo-7B & Qwen-7B & 0.51 & 0.65 & 0.13 & 0.636 & 0.429 & 0.466 & \textbf{0.843} & 0.621 & 0.431 & 0.488 & \textbf{0.723} \\
OLMo-7B & Mistral-7B & 0.51 & 0.64 & 0.12 & 0.636 & 0.429 & 0.588 & \textbf{0.884} & 0.659 & 0.406 & 0.626 & \textbf{0.724} \\
Llama-3-8B & OLMo-7B & 0.43 & 0.51 & 0.08 & 0.859 & \textbf{0.885} & 0.587 & 0.824 & 1.089 & \textbf{1.113} & 0.743 & 1.105 \\
Mistral-7B & Qwen-7B$^{\dagger}$ & 0.64 & 0.65 & 0.01 & \textbf{0.847} & 0.301 & 0.421 & 0.766 & \textbf{2.179} & -1.093 & -0.574 & 0.086 \\
\midrule
\multicolumn{13}{l}{\textit{Cross-family asymmetric}} \\
Gemma-270M & Qwen-7B & 0.22 & 0.65 & 0.42 & 0.604 & 0.598 & \textbf{0.777} & 0.678 & 0.523 & 0.525 & \textbf{0.634} & 0.492 \\
Gemma-270M & Llama-1B & 0.22 & 0.55 & 0.32 & 0.604 & 0.598 & \textbf{0.687} & 0.682 & 0.507 & 0.508 & \textbf{0.594} & 0.445 \\
Gemma-270M & Llama-3-8B & 0.22 & 0.43 & 0.21 & 0.604 & 0.598 & \textbf{0.769} & 0.628 & 0.525 & 0.537 & \textbf{0.666} & 0.463 \\
Gemma-270M & Qwen-0.5B & 0.22 & 0.41 & 0.18 & 0.604 & 0.598 & \textbf{0.653} & 0.604 & 0.486 & 0.496 & \textbf{0.589} & 0.373 \\
Qwen-0.5B & Llama-1B & 0.41 & 0.55 & 0.14 & \textbf{0.778} & 0.730 & 0.520 & 0.554 & \textbf{0.636} & 0.628 & 0.518 & 0.415 \\
Llama-1B & Qwen-7B & 0.55 & 0.65 & 0.10 & 0.804 & \textbf{0.810} & 0.688 & 0.789 & 0.691 & 0.720 & \textbf{0.761} & 0.574 \\
Qwen-0.5B & Llama-3-8B$^{\dagger}$ & 0.41 & 0.43 & 0.03 & \textbf{0.778} & 0.730 & 0.616 & 0.521 & \textbf{1.395} & 1.270 & 1.352 & 0.457 \\
\midrule
\multicolumn{13}{l}{\textit{Same-family asymmetric}} \\
Qwen-0.5B & Qwen-7B & 0.41 & 0.65 & 0.24 & 0.778 & 0.730 & 0.816 & \textbf{0.825} & 0.621 & 0.604 & \textbf{0.738} & 0.610 \\
\midrule
\textit{Mean (all, $n{=}14$)} &  &  &  &  & 0.732 & 0.658 & 0.656 & \textbf{0.743} & \textbf{0.825} & 0.556 & 0.628 & 0.573 \\
\textit{Mean (gap${>}0.05$, $n{=}12$)} &  &  &  &  & 0.719 & 0.681 & 0.679 & \textbf{0.759} & 0.665 & 0.634 & \textbf{0.668} & 0.623 \\
\bottomrule
\end{tabular}
\end{table*}

\vspace{2em}

\begin{table*}[h]
\centering
\scriptsize
\setlength{\tabcolsep}{3pt}
\caption{Full results on TriviaQA (no context). Pairs grouped by type and sorted by capability gap (descending). $^{\dagger}$ marks rows with $\text{gap} \le 0.05$. \textbf{Bold} = best per row among G-Ent, G-PPL, CMP, CME.}
\label{tab:triviaqa-noctx}
\begin{tabular}{llccc | cccc | cccc}
\toprule
& & & & & \multicolumn{4}{c|}{AUROC $\uparrow$} & \multicolumn{4}{c}{APGR $\uparrow$} \\
Generator & Verifier & Acc$_g$ & Acc$_v$ & Gap & G-Ent & G-PPL & CMP & CME & G-Ent & G-PPL & CMP & CME \\
\midrule
\multicolumn{13}{l}{\textit{Same-size cross-family}} \\
OLMo-7B & Llama-3-8B & 0.39 & 0.67 & 0.29 & 0.774 & 0.504 & 0.706 & \textbf{0.807} & \textbf{0.664} & 0.484 & 0.628 & 0.601 \\
OLMo-7B & Mistral-7B & 0.39 & 0.66 & 0.28 & 0.774 & 0.504 & 0.812 & \textbf{0.883} & 0.659 & 0.491 & \textbf{0.729} & 0.634 \\
OLMo-7B & Qwen-7B & 0.39 & 0.59 & 0.21 & 0.774 & 0.504 & 0.673 & \textbf{0.849} & \textbf{0.635} & 0.480 & 0.616 & 0.596 \\
Qwen-7B & Llama-3-8B & 0.59 & 0.67 & 0.08 & \textbf{0.823} & 0.793 & 0.748 & 0.625 & \textbf{0.980} & 0.946 & 0.964 & 0.533 \\
Qwen-7B & Mistral-7B & 0.59 & 0.66 & 0.07 & \textbf{0.823} & 0.793 & 0.703 & 0.670 & 0.859 & 0.837 & \textbf{0.880} & 0.572 \\
Mistral-7B & Llama-3-8B$^{\dagger}$ & 0.66 & 0.67 & 0.01 & \textbf{0.851} & 0.483 & 0.597 & 0.747 & \textbf{3.438} & -0.486 & 0.812 & 0.410 \\
\midrule
\multicolumn{13}{l}{\textit{Cross-family asymmetric}} \\
Gemma-270M & Llama-3-8B & 0.10 & 0.67 & 0.57 & 0.710 & 0.672 & \textbf{0.836} & 0.698 & 0.526 & 0.527 & \textbf{0.573} & 0.469 \\
Gemma-270M & Qwen-7B & 0.10 & 0.59 & 0.49 & 0.710 & 0.672 & \textbf{0.796} & 0.683 & 0.504 & 0.507 & \textbf{0.594} & 0.420 \\
Qwen-0.5B & Llama-3-8B & 0.21 & 0.67 & 0.46 & \textbf{0.782} & 0.775 & 0.776 & 0.587 & 0.557 & 0.555 & \textbf{0.614} & 0.493 \\
Gemma-270M & Llama-1B & 0.10 & 0.42 & 0.32 & 0.710 & 0.672 & \textbf{0.773} & 0.739 & 0.488 & 0.500 & \textbf{0.572} & 0.375 \\
Qwen-0.5B & Llama-1B & 0.21 & 0.42 & 0.21 & \textbf{0.782} & 0.775 & 0.623 & 0.658 & 0.533 & 0.523 & \textbf{0.593} & 0.432 \\
Llama-1B & Qwen-7B & 0.42 & 0.59 & 0.17 & 0.840 & \textbf{0.851} & 0.738 & 0.806 & 0.653 & 0.671 & \textbf{0.759} & 0.501 \\
Gemma-270M & Qwen-0.5B & 0.10 & 0.21 & 0.11 & 0.710 & 0.672 & \textbf{0.737} & 0.686 & 0.493 & 0.521 & \textbf{0.555} & 0.293 \\
\midrule
\multicolumn{13}{l}{\textit{Same-family asymmetric}} \\
Qwen-0.5B & Qwen-7B & 0.21 & 0.59 & 0.38 & 0.782 & 0.775 & 0.793 & \textbf{0.817} & 0.530 & 0.532 & \textbf{0.644} & 0.510 \\
Llama-1B & Llama-3-8B & 0.42 & 0.67 & 0.25 & 0.840 & 0.851 & \textbf{0.938} & 0.814 & 0.685 & 0.693 & \textbf{0.811} & 0.589 \\
\midrule
\textit{Mean (all, $n{=}15$)} &  &  &  &  & \textbf{0.779} & 0.687 & 0.750 & 0.738 & \textbf{0.814} & 0.519 & 0.690 & 0.495 \\
\textit{Mean (gap${>}0.05$, $n{=}14$)} &  &  &  &  & \textbf{0.774} & 0.701 & 0.761 & 0.737 & 0.626 & 0.590 & \textbf{0.681} & 0.501 \\
\bottomrule
\end{tabular}
\end{table*}

\vspace{2em}

\begin{table*}[h]
\centering
\scriptsize
\setlength{\tabcolsep}{3pt}
\caption{Full results on GSM8K. Pairs grouped by type and sorted by capability gap (descending). No row has $\text{gap} \le 0.05$, so the filtered mean coincides with the full mean. \textbf{Bold} = best per row among G-Ent, G-PPL, CMP, CME.}
\label{tab:gsm8k}
\begin{tabular}{llccc | cccc | cccc}
\toprule
& & & & & \multicolumn{4}{c|}{AUROC $\uparrow$} & \multicolumn{4}{c}{APGR $\uparrow$} \\
Generator & Verifier & Acc$_g$ & Acc$_v$ & Gap & G-Ent & G-PPL & CMP & CME & G-Ent & G-PPL & CMP & CME \\
\midrule
\multicolumn{13}{l}{\textit{Same-size cross-family}} \\
Mistral-7B & Qwen-7B & 0.54 & 0.90 & 0.36 & 0.658 & 0.542 & 0.662 & \textbf{0.718} & 0.583 & 0.522 & 0.606 & \textbf{0.620} \\
Mistral-7B & OLMo-7B & 0.54 & 0.84 & 0.30 & 0.658 & 0.542 & \textbf{0.722} & 0.684 & 0.581 & 0.523 & \textbf{0.634} & 0.592 \\
Llama-3-8B & Qwen-7B & 0.61 & 0.90 & 0.29 & 0.593 & 0.577 & \textbf{0.609} & 0.508 & 0.549 & 0.534 & \textbf{0.594} & 0.492 \\
Llama-3-8B & OLMo-7B & 0.61 & 0.84 & 0.22 & 0.593 & 0.577 & \textbf{0.633} & 0.527 & 0.530 & 0.512 & \textbf{0.592} & 0.469 \\
Mistral-7B & Llama-3-8B & 0.54 & 0.61 & 0.07 & 0.658 & 0.542 & \textbf{0.704} & 0.675 & 0.955 & 0.663 & \textbf{1.155} & 0.977 \\
OLMo-7B & Qwen-7B & 0.84 & 0.90 & 0.07 & 0.751 & 0.555 & 0.739 & \textbf{0.774} & 0.881 & 0.683 & \textbf{0.904} & 0.871 \\
\midrule
\multicolumn{13}{l}{\textit{Cross-family asymmetric}} \\
Gemma-270M & Qwen-7B & 0.05 & 0.90 & 0.85 & 0.453 & 0.425 & 0.530 & \textbf{0.580} & 0.497 & 0.496 & \textbf{0.507} & 0.503 \\
Gemma-270M & Llama-3-8B & 0.05 & 0.61 & 0.57 & 0.453 & 0.425 & 0.530 & \textbf{0.561} & 0.487 & 0.487 & \textbf{0.505} & 0.503 \\
Llama-1B & Qwen-7B & 0.37 & 0.90 & 0.53 & 0.559 & 0.531 & 0.652 & \textbf{0.659} & 0.515 & 0.506 & \textbf{0.563} & 0.556 \\
Qwen-0.5B & Llama-3-8B & 0.24 & 0.61 & 0.37 & 0.640 & 0.630 & 0.644 & \textbf{0.652} & 0.542 & 0.539 & 0.543 & \textbf{0.546} \\
Gemma-270M & Llama-1B & 0.05 & 0.37 & 0.32 & 0.453 & 0.425 & 0.503 & \textbf{0.520} & 0.492 & 0.498 & \textbf{0.520} & 0.500 \\
Gemma-270M & Qwen-0.5B & 0.05 & 0.24 & 0.19 & 0.453 & 0.425 & 0.496 & \textbf{0.513} & 0.470 & 0.481 & \textbf{0.534} & 0.509 \\
Qwen-0.5B & Llama-1B & 0.24 & 0.37 & 0.13 & \textbf{0.640} & 0.630 & 0.608 & 0.627 & 0.597 & 0.591 & 0.597 & \textbf{0.601} \\
\midrule
\multicolumn{13}{l}{\textit{Same-family asymmetric}} \\
Qwen-0.5B & Qwen-7B & 0.24 & 0.90 & 0.66 & 0.640 & 0.630 & 0.664 & \textbf{0.698} & 0.527 & 0.528 & 0.537 & \textbf{0.543} \\
Llama-1B & Llama-3-8B & 0.37 & 0.61 & 0.25 & 0.559 & 0.531 & \textbf{0.653} & 0.614 & 0.536 & 0.517 & \textbf{0.633} & 0.589 \\
\midrule
\textit{Mean (all, $n{=}15$)} &  &  &  &  & 0.584 & 0.533 & \textbf{0.623} & 0.621 & 0.583 & 0.539 & \textbf{0.628} & 0.591 \\
\bottomrule
\end{tabular}
\end{table*}

\section{Baseline Comparison}

\begin{table}[t]
\centering
\small
\setlength{\tabcolsep}{4pt}
\begin{tabular}{lllrrrrrr}
\toprule
Dataset & Generator & Verifier & Gap & CMP & CME & P(True) & V-Agree & Sem-Ent (k=10) \\
\midrule
MMLU & Gemma-270M & Qwen-7B & +47.5 & \textbf{0.828} & 0.609 & 0.572 & 0.820 & 0.515 \\
 & Gemma-270M & Llama-3-8B & +37.4 & 0.710 & 0.597 & 0.503 & \textbf{0.739} & 0.515 \\
 & Qwen-0.5B & Qwen-7B & +29.9 & 0.841 & 0.583 & 0.551 & \textbf{0.843} & 0.633 \\
 & Llama-1B & Qwen-7B & +28.8 & 0.786 & 0.634 & 0.547 & \textbf{0.821} & 0.617 \\
 & Qwen-0.5B & Llama-3-8B & +19.9 & 0.749 & 0.639 & 0.498 & \textbf{0.784} & 0.633 \\
 & Llama-1B & Llama-3-8B & +18.8 & \textbf{0.817} & 0.679 & 0.475 & 0.748 & 0.617 \\
 & Gemma-270M & Llama-1B & +18.6 & 0.564 & 0.532 & 0.489 & \textbf{0.612} & 0.515 \\
 & Gemma-270M & Qwen-0.5B & +17.5 & 0.566 & 0.526 & 0.515 & \textbf{0.614} & 0.515 \\
 & Mistral-7B & Qwen-7B & +15.8 & 0.711 & 0.632 & 0.551 & \textbf{0.800} & 0.627 \\
 & OLMo-7B & Qwen-7B & +15.4 & \textbf{0.896} & 0.697 & 0.546 & 0.810 & 0.695 \\
 & Llama-3-8B & Qwen-7B & +10.1 & 0.694 & 0.635 & 0.508 & \textbf{0.760} & 0.744 \\
 & Mistral-7B & Llama-3-8B & +5.8 & \textbf{0.753} & 0.716 & 0.494 & 0.737 & 0.627 \\
 & OLMo-7B & Llama-3-8B & +5.3 & \textbf{0.822} & 0.689 & 0.485 & 0.738 & 0.695 \\
\midrule
TriviaQA & Gemma-270M & Qwen-7B & +42.4 & 0.777 & 0.678 & 0.590 & \textbf{0.826} & 0.716 \\
 & Gemma-270M & Llama-1B & +32.3 & 0.687 & 0.682 & 0.433 & \textbf{0.809} & 0.716 \\
 & Qwen-0.5B & Qwen-7B & +23.8 & 0.816 & \textbf{0.825} & 0.595 & 0.775 & 0.783 \\
 & Gemma-270M & Llama-3-8B & +21.2 & \textbf{0.769} & 0.628 & 0.431 & 0.694 & 0.716 \\
 & Llama-3-8B & Qwen-7B & +21.1 & 0.772 & \textbf{0.875} & 0.540 & 0.792 & 0.700 \\
 & Llama-3-8B & Mistral-7B & +20.2 & 0.825 & \textbf{0.922} & 0.476 & 0.764 & 0.700 \\
 & Gemma-270M & Qwen-0.5B & +18.5 & 0.653 & 0.604 & 0.432 & 0.692 & \textbf{0.716} \\
 & Qwen-0.5B & Llama-1B & +13.8 & 0.520 & 0.554 & 0.452 & 0.745 & \textbf{0.783} \\
 & OLMo-7B & Qwen-7B & +13.4 & 0.466 & 0.843 & 0.582 & \textbf{0.849} & 0.797 \\
 & OLMo-7B & Mistral-7B & +12.4 & 0.588 & \textbf{0.884} & 0.519 & 0.837 & 0.797 \\
 & Llama-1B & Qwen-7B & +10.1 & 0.688 & 0.789 & 0.591 & 0.835 & \textbf{0.838} \\
 & Llama-3-8B & OLMo-7B & +7.8 & 0.587 & \textbf{0.824} & 0.478 & 0.763 & 0.700 \\
\midrule
TriviaQA (no ctx) & Gemma-270M & Llama-3-8B & +56.7 & \textbf{0.836} & 0.698 & 0.458 & 0.712 & 0.720 \\
 & Gemma-270M & Qwen-7B & +49.0 & \textbf{0.796} & 0.683 & 0.577 & 0.726 & 0.720 \\
 & Qwen-0.5B & Llama-3-8B & +45.6 & 0.776 & 0.587 & 0.500 & 0.802 & \textbf{0.824} \\
 & Qwen-0.5B & Qwen-7B & +38.0 & 0.793 & 0.817 & 0.651 & \textbf{0.833} & 0.824 \\
 & Gemma-270M & Llama-1B & +31.6 & \textbf{0.773} & 0.739 & 0.445 & 0.711 & 0.720 \\
 & OLMo-7B & Llama-3-8B & +28.5 & 0.706 & 0.807 & 0.501 & 0.833 & \textbf{0.846} \\
 & OLMo-7B & Mistral-7B & +27.8 & 0.812 & \textbf{0.883} & 0.576 & 0.838 & 0.846 \\
 & Llama-1B & Llama-3-8B & +25.1 & \textbf{0.938} & 0.814 & 0.500 & 0.845 & 0.837 \\
 & OLMo-7B & Qwen-7B & +20.8 & 0.673 & \textbf{0.849} & 0.619 & 0.827 & 0.846 \\
 & Qwen-0.5B & Llama-1B & +20.6 & 0.623 & 0.658 & 0.478 & 0.787 & \textbf{0.824} \\
 & Llama-1B & Qwen-7B & +17.3 & 0.738 & 0.806 & 0.604 & \textbf{0.839} & 0.837 \\
 & Gemma-270M & Qwen-0.5B & +11.1 & \textbf{0.737} & 0.686 & 0.438 & 0.673 & 0.720 \\
 & Qwen-7B & Llama-3-8B & +7.7 & 0.748 & 0.625 & 0.482 & 0.809 & \textbf{0.845} \\
 & Qwen-7B & Mistral-7B & +6.9 & 0.703 & 0.670 & 0.533 & 0.816 & \textbf{0.845} \\
\midrule
GSM8K & Gemma-270M & Qwen-7B & +85.4 & 0.530 & 0.580 & 0.745 & \textbf{0.956} & 0.610 \\
 & Qwen-0.5B & Qwen-7B & +66.3 & 0.664 & 0.698 & 0.628 & \textbf{0.970} & 0.659 \\
 & Gemma-270M & Llama-3-8B & +56.6 & 0.530 & 0.561 & 0.702 & \textbf{0.846} & 0.610 \\
 & Llama-1B & Qwen-7B & +53.4 & 0.652 & 0.659 & 0.643 & \textbf{0.985} & 0.748 \\
 & Qwen-0.5B & Llama-3-8B & +37.5 & 0.644 & 0.652 & 0.664 & \textbf{0.910} & 0.659 \\
 & Mistral-7B & Qwen-7B & +36.2 & 0.662 & 0.718 & 0.635 & \textbf{0.963} & 0.828 \\
 & Gemma-270M & Llama-1B & +32.0 & 0.503 & 0.520 & 0.663 & \textbf{0.883} & 0.610 \\
 & Mistral-7B & OLMo-7B & +29.6 & 0.722 & 0.684 & 0.481 & \textbf{0.949} & 0.828 \\
 & Llama-3-8B & Qwen-7B & +28.9 & 0.609 & 0.508 & 0.583 & \textbf{0.950} & 0.821 \\
 & Llama-1B & Llama-3-8B & +24.6 & 0.653 & 0.614 & 0.687 & \textbf{0.875} & 0.748 \\
 & Llama-3-8B & OLMo-7B & +22.4 & 0.633 & 0.527 & 0.416 & \textbf{0.922} & 0.821 \\
 & Gemma-270M & Qwen-0.5B & +19.1 & 0.496 & 0.513 & 0.535 & \textbf{0.737} & 0.610 \\
 & Qwen-0.5B & Llama-1B & +12.9 & 0.608 & 0.627 & 0.503 & \textbf{0.792} & 0.659 \\
 & Mistral-7B & Llama-3-8B & +7.3 & 0.704 & 0.675 & 0.607 & 0.815 & \textbf{0.828} \\
 & OLMo-7B & Qwen-7B & +6.5 & 0.739 & 0.774 & 0.554 & \textbf{0.916} & 0.879 \\
\bottomrule
\end{tabular}
\caption{AUROC by correctness signal across (generator, verifier, dataset) tuples with gap $>$ 5pp. CMP and CME are this paper's cross-model perplexity / entropy signals (single prefill on the verifier). P(True), Verifier Agreement, and Semantic Entropy \cite{kuhn2023semantic} are external baselines. \textbf{Bold} = best AUROC per row.}
\label{tab:combined_auroc}
\end{table}
\subsection{Selective Prediction Analysis}
\label{sec:selective-prediction}

The aggregate metrics in Section~4 (AUROC, APGR, quintile spread)
summarise each signal's ability to rank instances by expected
correctness. A complementary perspective comes from
\emph{selective prediction}~\citep{geifman2017selective}, which
asks: if we are allowed to abstain on a fraction of inputs, how
accurate can we be on the remainder?

We construct coverage--accuracy curves by sweeping an abstention
threshold on each signal score. At each threshold we retain only
the instances on which the signal is most confident (lowest score)
and compute generator accuracy on that retained subset. A stronger
signal produces a curve that lies higher: it achieves greater
accuracy at the same coverage, or equivalently reaches a target
accuracy while retaining more examples.

\begin{figure}[t]
  \centering
  \includegraphics[width=\linewidth]{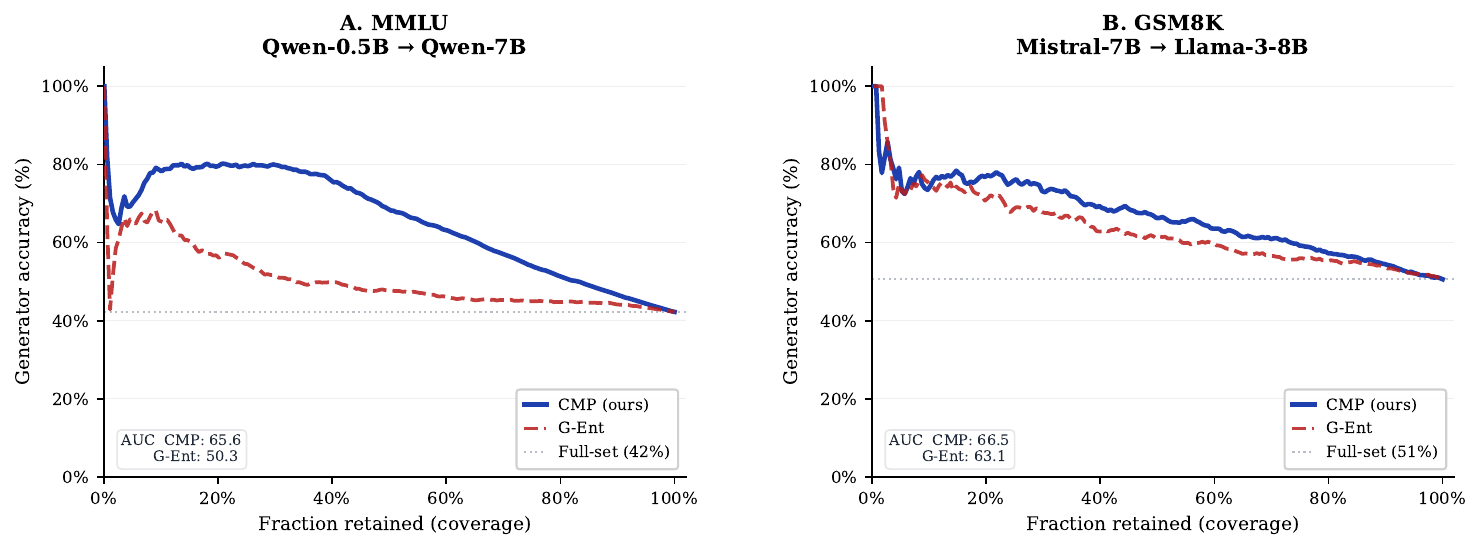}
  \caption{%
    \textbf{Coverage--accuracy curves (single pair).}
    At each coverage level we abstain on instances whose signal
    score exceeds a swept threshold and report generator accuracy
    on the retained subset. CMP (blue, solid) maintains higher
    accuracy than G-Ent (red, dashed) across nearly all operating
    points on both MMLU (Qwen-0.5B $\to$ Qwen-7B) and GSM8K
    (Mistral-7B $\to$ Llama-3-8B). The dotted line marks
    full-set accuracy.}
  \label{fig:coverage-accuracy-single}
\end{figure}

\paragraph{Single-pair examples.}
Figure~\ref{fig:coverage-accuracy-single} shows curves for two
representative pairs (the same ones featured in the quintile
figure): Qwen-0.5B $\to$ Qwen-7B on MMLU and Mistral-7B $\to$
Llama-3-8B on GSM8K. On MMLU, CMP dominates G-Ent across nearly
the full coverage range; at 20\% coverage the most confident CMP
predictions approach perfect accuracy, whereas G-Ent remains
closer to the full-set baseline. On GSM8K the two curves are
closer together, but CMP maintains a consistent advantage,
particularly at low coverage.

\begin{figure}[t]
  \centering
  \includegraphics[width=\linewidth]{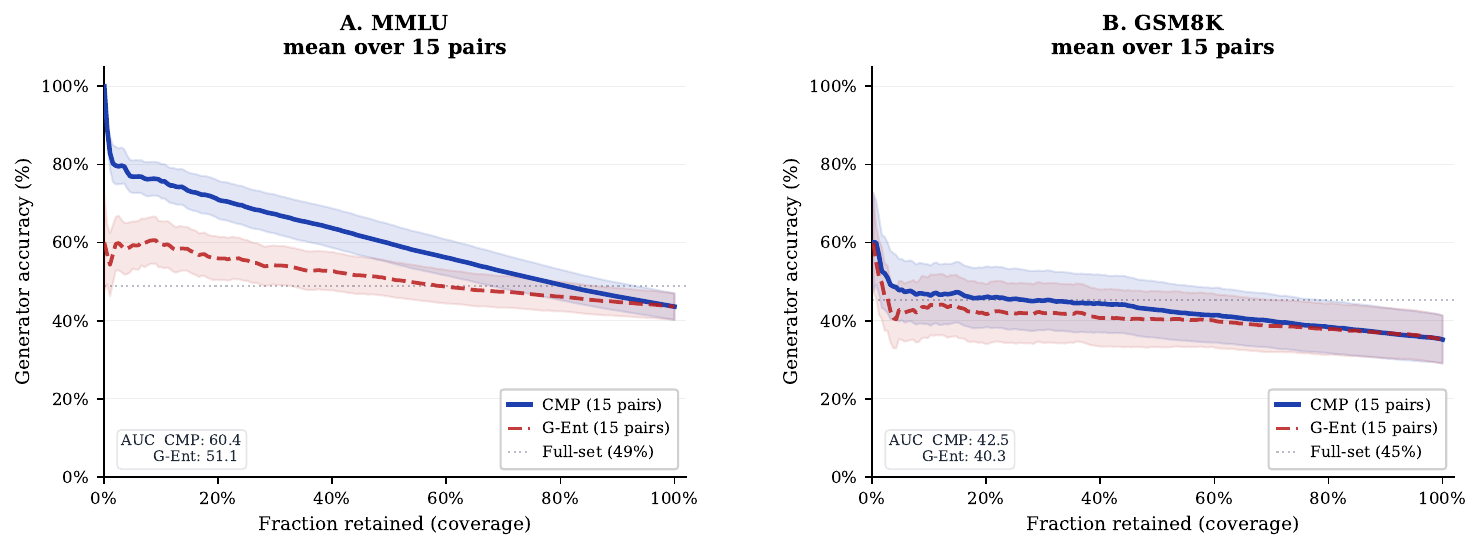}
  \caption{%
    \textbf{Coverage--accuracy curves (averaged over all pairs).}
    Curves are averaged across all 15 model pairs per benchmark;
    shaded bands show $\pm$1 SE. CMP consistently achieves higher
    accuracy than G-Ent at equal coverage on both MMLU and GSM8K.}
  \label{fig:coverage-accuracy-avg}
\end{figure}

\paragraph{Aggregate trends.}
To verify that these patterns are not pair-specific,
Figure~\ref{fig:coverage-accuracy-avg} averages the
coverage--accuracy curves across all 15 model pairs per benchmark
(shaded bands show $\pm$1 standard error). The averaged curves
confirm the single-pair findings: CMP lies above G-Ent at
virtually every operating point on both MMLU (AUC 60.4 vs.\ 53.1)
and GSM8K (AUC 42.5 vs.\ 40.3). These AUC summaries are
consistent with the AUROC and quintile-spread rankings in
Section~4, and confirm that the advantage of cross-model
perplexity extends to the operationally relevant setting where a
system must decide, per query, whether to serve a generation or
abstain. This makes CMP directly applicable as an abstention trigger in high-stakes settings where a confident wrong answer is costlier than no answer.

\section{Effect of Capability Gap}

\begin{figure*}[h]
  \centering
  \includegraphics[width=\linewidth]{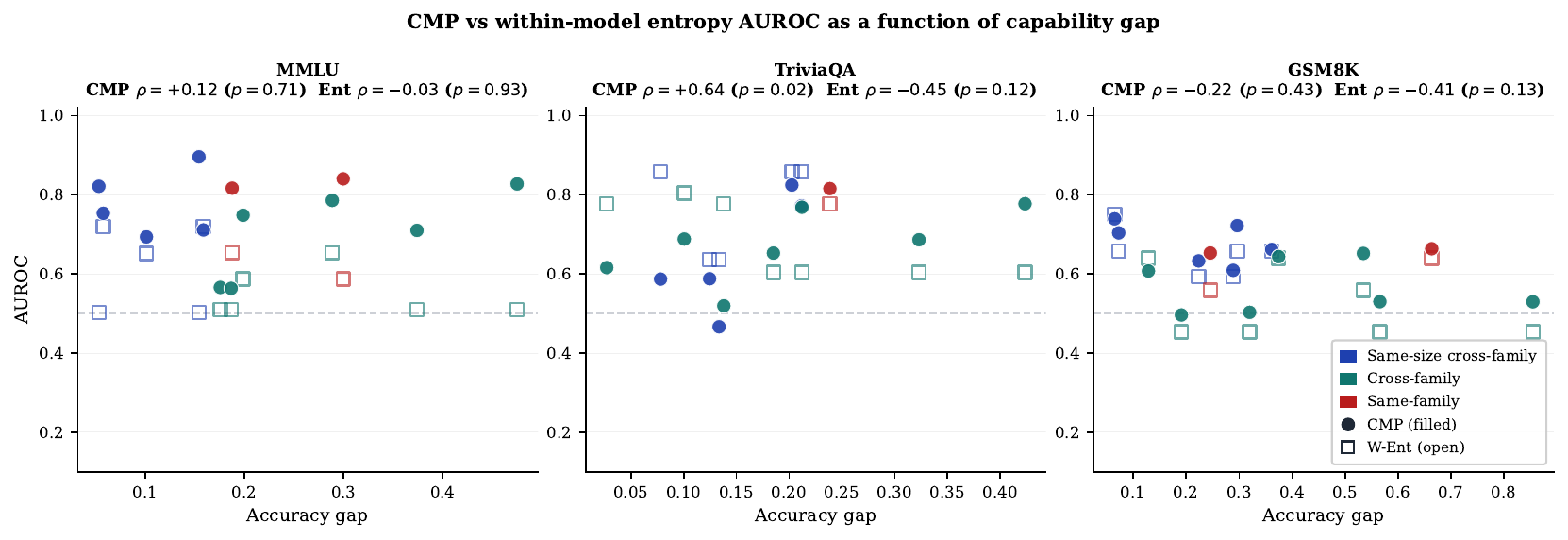}
  \caption{\Gls{cmp} (filled circles) and within-model entropy (open squares) AUROC
  versus capability gap. The positive gap effect on TriviaQA is specific to \gls{cmp}
  --- entropy AUROC is uncorrelated with gap on all three datasets.}
  \label{fig:gap_v1}
\end{figure*}

\begin{figure*}[h]
  \centering
  \includegraphics[width=\linewidth]{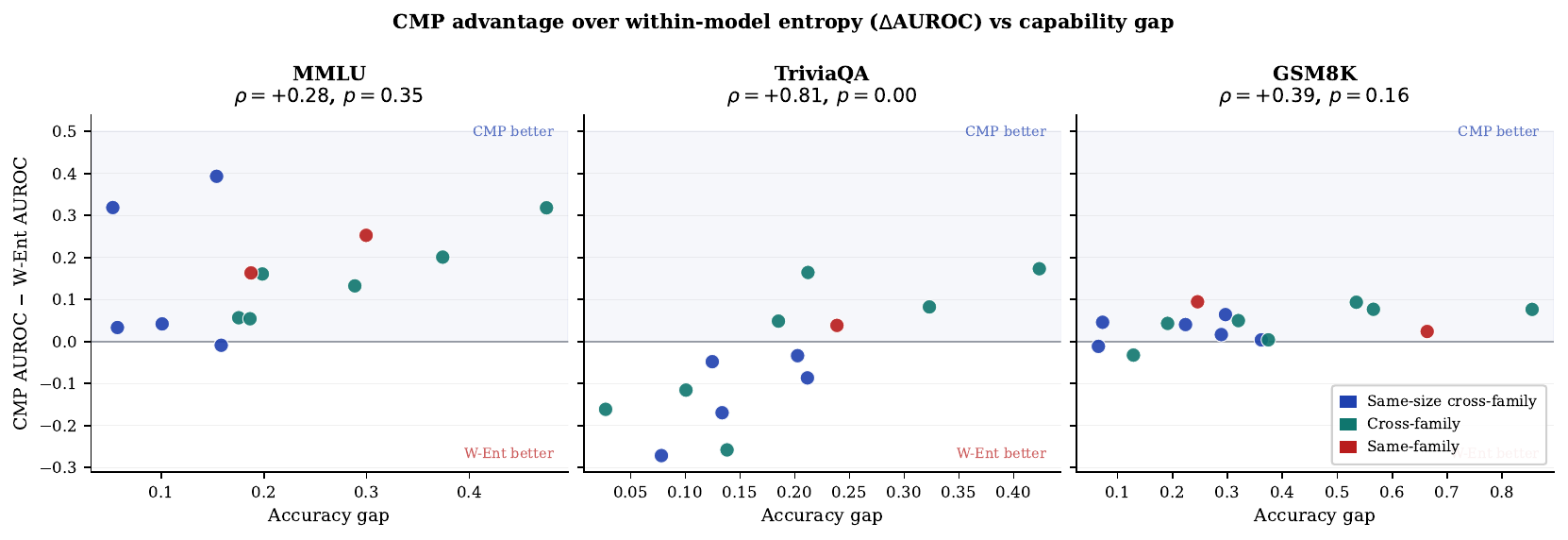}
  \caption{\Gls{cmp} advantage over within-model entropy ($\Delta$AUROC = CMP $-$
  G-Ent) versus capability gap. \Gls{cmp}'s advantage grows significantly with gap
  on TriviaQA ($\rho=+0.81$, $p<0.01$), while on MMLU \gls{cmp} leads regardless
  of gap size. GSM8K shows a positive but non-significant trend ($\rho=+0.39$,
  $p=0.16$).}
  \label{fig:gap_v2}
\end{figure*}

Figures~\ref{fig:gap_v1} and~\ref{fig:gap_v2} provide
additional context for the capability gap by including within-model entropy as a comparison signal.
Figure~\ref{fig:gap_v1} plots both \gls{cmp} and entropy AUROC on the same axes,
showing that the positive gap effect on TriviaQA is specific to \gls{cmp} --- entropy
shows no significant correlation with gap on any dataset. Figure~\ref{fig:gap_v2}
plots the difference directly ($\Delta$AUROC = CMP $-$ G-Ent), confirming that
\gls{cmp}'s advantage over entropy grows with capability gap on TriviaQA
($\rho=+0.81$, $p<0.01$) but not on MMLU or GSM8K.

\clearpage

\section{Experimental Setup Details}
\label{app:setup}

\textbf{Models.}
We evaluate across a diverse set of model pairs spanning multiple
families and capability levels. Generating models include
Qwen2.5-0.5B-Instruct, Qwen2.5-1.5B-Instruct, Llama-3.2-1B-Instruct,
Llama-3.2-3B-Instruct, gemma-3-270m-it, and SmolLM2-1.7B-Instruct.
Verifying models include Qwen2.5-7B-Instruct and
Meta-Llama-3-8B-Instruct. We additionally evaluate
\textit{same-sized cross-family} pairs---models with comparable
parameter counts but different architectures and training corpora,
including Mistral-7B-Instruct-v0.3, OLMo-3-7B-Instruct, and
Meta-Llama-3-8B-Instruct as both generators and verifiers---to isolate
the effect of model diversity from capability asymmetry.

\textbf{Datasets.}
We evaluate on four benchmarks spanning distinct task types.
\textbf{MMLU} \citep{hendrycks2021measuring} is a 57-subject
multiple-choice benchmark testing knowledge and reasoning; we use
single-token answer generation. \textbf{TriviaQA}
\citep{joshi2017triviaqa} is evaluated in both the no-context variant,
testing pure knowledge retrieval, and the context variant, testing
reading comprehension. \textbf{GSM8K} \citep{cobbe2021gsm8k} requires
multi-step chain-of-thought reasoning, with answers evaluated by
extracting the final numerical value. We evaluate on up to 2000
examples per dataset with a fixed random seed.

\textbf{Implementation.}
All models are loaded in bfloat16 precision. Generations use greedy
decoding with maximum token lengths of 5 (MMLU), 8 (TriviaQA), and
256 (GSM8K). The verifying model performs only a single prefill forward
pass and never generates tokens. All experiments are run on a single
NVIDIA A10G GPU (24GB).

\section{Relationship to Generation-Based Correctness Signals}
\label{app:baselines}

The baselines in our main experiments---generator entropy (G-Ent) and generator 
perplexity (G-PPL)---are chosen as direct ablations of the cross-model design 
decision: they answer whether the verifying model's distribution adds signal over 
the generating model's own distribution, holding the computational regime constant. 
This appendix clarifies how CMP and CME relate to generation-based correctness 
signals that occupy a different point on the compute-accuracy tradeoff.

\subsection*{D.1 Consistency-Based Methods}

SelfCheckGPT \citep{manakul2023selfcheckgpt} and CrossCheckGPT 
\citep{sun2024crosscheckgpt} detect errors by measuring consistency across 
multiple stochastic samples from the same model or across outputs from independent 
models. These methods can achieve strong hallucination detection performance, but 
operate in a fundamentally different compute regime: they require $k$ full 
autoregressive generation passes, where $k$ is typically 3--5, compared to the 
single prefill forward pass required by CMP and CME. For answer lengths typical 
of our benchmarks (up to 256 tokens for GSM8K) this represents a substantial 
difference in inference cost.

Beyond compute, consistency-based methods are conceptually distinct from 
cross-model disagreement. SelfCheckGPT measures whether a model agrees with 
itself across samples; CMP measures whether a second model is surprised by the 
generating model's specific answer. The two signals are complementary: 
consistency captures within-model variance, while CMP captures 
between-model disagreement on a fixed output. CMP could in principle be extended 
to sampled rather than greedy answers---computing mean verifier perplexity over 
$k$ samples---which would interpolate between the two approaches and may improve 
signal quality at higher compute cost. We leave this extension to future work.

We do not claim that CMP outperforms consistency-based methods unconditionally. 
Rather, CMP occupies the low-cost end of the correctness signal spectrum: it 
provides a meaningful signal with no labeled data, no generation from the 
verifier, and no repeated sampling from the generator. This makes it most 
attractive in latency-constrained or cost-constrained deployment settings where 
multiple generation passes are not feasible.

\subsection*{D.2 Generation-Based Judge Methods}

A related line of work uses a language model as a judge to evaluate another 
model's outputs \citep{zheng2023judging, kim2024prometheus, dubois2024alpacafarm}. 
These approaches require the judge model to \emph{generate} a response---typically 
a scalar rating or structured critique---which is one to two orders of magnitude 
more expensive than a prefill pass. They also target output quality and preference 
rather than per-instance binary correctness, making direct AUROC comparison 
inappropriate. Self-evaluation methods such as P(True) \citep{kadavath2022language} 
similarly require generation from the evaluating model and are known to be 
unreliable on smaller models, which constitute most of our generator set.

CMP and CME differ from all of these in that the verifying model never generates 
tokens: it performs only a prefill pass and produces a scalar signal directly 
from its output logits. This constraint is what enables the single-forward-pass 
efficiency claim, and it is what distinguishes our contribution from the 
LLM-as-Judge literature.

\subsection*{D.3 Why G-Ent and G-PPL Are the Appropriate Ablations}

Given the above, G-Ent and G-PPL are the correct ablation baselines for our 
research question. The central claim of this paper is that a verifying model's 
distribution over a generating model's answer contains correctness signal beyond 
what is already present in the generating model's own distribution. G-Ent and 
G-PPL isolate exactly this question: they are the within-model analogues of CME 
and CMP respectively, computed from the same answer under the same greedy decoding 
setting, with no verifier involved. Any performance gap between CMP and G-PPL, 
or between CME and G-Ent, is attributable solely to the cross-model signal. 
Comparing against generation-based methods would conflate this signal with 
sampling variance, generation length, or judge calibration, obscuring the 
contribution we seek to measure.
\section{Failure Mode: Very Weak Generators on Reasoning Tasks}
\label{sec:weak-generators}

CMP requires the generator to produce a meaningful mixture of correct and 
incorrect answers. On knowledge tasks (MMLU, TriviaQA), even Gemma-3-270M 
at 10--24\% accuracy provides enough signal for CMP to achieve AUROC 
0.56--0.84. However, on GSM8K where Gemma-3-270M achieves only 5\% 
accuracy. 
At this accuracy level, virtually all generated answers are incorrect 
chain-of-thought reasoning with wrong final numbers, leaving the verifier 
no distributional contrast to exploit. We observe that CMP degrades 
gracefully: the threshold for useful signal is approximately 10\% 
generator accuracy, below which both CMP and generator-entropy baselines 
become uninformative.

\begin{tabular}{lccccccc}
\toprule
& & & \multicolumn{2}{c}{AUROC $\uparrow$} & \multicolumn{2}{c}{APGR $\uparrow$} \\
\cmidrule(lr){4-5} \cmidrule(lr){6-7}
Verifier & Acc$_v$ & Gap & G-Ent & CMP & G-Ent & CMP \\
\midrule
Qwen-7B & 0.90 & 0.85 & 0.453 & 0.530 & 0.497 & 0.507 \\
Llama-3-8B & 0.61 & 0.57 & 0.453 & 0.530 & 0.487 & 0.505 \\
Llama-1B & 0.37 & 0.32 & 0.453 & 0.503 & 0.492 & 0.520 \\
Qwen-0.5B & 0.24 & 0.19 & 0.453 & 0.496 & 0.470 & 0.534 \\
\midrule
Mean & & & 0.453 & 0.515 & 0.487 & 0.516 \\
\bottomrule
\label{tab:gemma-gsm8k}
\end{tabular}

\section{GSM8K: Final-Answer Tokens and Verification Prompting}
\label{sec:gsm8k-ablation}

The chain-of-thought sequences in GSM8K raise a natural question:
does the routing signal come from the full reasoning trace or only
the final numerical answer? We additionally test whether an explicit
verification prompt---asking the verifier ``Is this answer
correct?''---can match or exceed the implicit signal in CMP.

\paragraph{Final-answer CMP.}
We extract the token positions corresponding to the final numerical
answer (the digits after the \texttt{ANSWER:} marker) and compute
CMP restricted to those positions only (CMP-Final), compared to CMP
over the full chain-of-thought (CMP-Full). Across 8 model pairs
(Table~\ref{tab:gsm8k-final}), CMP-Final achieves mean AUROC of
0.618 versus 0.663 for CMP-Full and mean APGR of 0.665 versus 0.682
(Table~\ref{tab:gsm8k-final-apgr}). On most pairs CMP-Full leads,
but CMP-Final matches or exceeds it on two pairs
(Llama-1B $\to$ Qwen-7B and OLMo-7B $\to$ Qwen-7B), suggesting
that the verifier's disagreement signal is largely concentrated
in the final answer rather than distributed across intermediate
reasoning steps. This implies CMP could be applied more efficiently
by scoring only the answer tokens, avoiding the cost of processing
the full chain-of-thought through the verifier.

\paragraph{Verification prompting.}
We test an explicit P(True) baseline \citep{kadavath2022language}:
we prompt the verifier with the question and the generator's
proposed answer, asking ``Is the proposed answer correct? Yes or
No,'' and extract the probability assigned to the ``Yes'' token.
Despite being a natural approach to answer verification, P(True)
achieves mean AUROC of only 0.463---\emph{below} random
(0.504)---and mean APGR of 0.492. The verifier's explicit judgment
of correctness is far less informative than its implicit
token-level perplexity on the same answer. This is consistent with
known limitations of LLM self-evaluation on mathematical reasoning:
models that cannot reliably solve a problem also cannot reliably
judge whether a solution is correct when asked directly.

\begin{table}[t]
\centering
\small
\setlength{\tabcolsep}{4pt}
\caption{GSM8K correctness prediction (AUROC): Full chain-of-thought CMP vs.\ final-answer-only CMP vs.\ P(True) verification prompting. CMP-Full and CMP-Final perform comparably, while P(True) is near random. \textbf{Bold} = best among CMP-Full, CMP-Final, P(True).}
\label{tab:gsm8k-final}
\begin{tabular}{llcc | ccc | cc}
\toprule
& & & & \multicolumn{3}{c|}{AUROC $\uparrow$} & \multicolumn{2}{c}{Baselines} \\
Generator & Verifier & Acc$_g$ & Gap & CMP-Full & CMP-Final & P(True) & G-Ent & G-PPL \\
\midrule
Qwen-0.5B & Qwen-7B & 0.25 & 0.64 & \textbf{0.629} & 0.456 & 0.495 & 0.616 & 0.621 \\
Llama-1B & Qwen-7B & 0.38 & 0.51 & 0.668 & \textbf{0.755} & 0.463 & 0.543 & 0.523 \\
Mistral-7B & Qwen-7B & 0.51 & 0.39 & \textbf{0.668} & 0.574 & 0.487 & 0.655 & 0.539 \\
Qwen-0.5B & Llama-3-8B & 0.25 & 0.36 & \textbf{0.627} & 0.525 & 0.431 & 0.616 & 0.621 \\
Llama-3-8B & Qwen-7B & 0.61 & 0.28 & \textbf{0.590} & 0.584 & 0.476 & 0.596 & 0.577 \\
Llama-1B & Llama-3-8B & 0.38 & 0.23 & \textbf{0.655} & 0.611 & 0.417 & 0.543 & 0.523 \\
Mistral-7B & Llama-3-8B & 0.51 & 0.11 & \textbf{0.719} & 0.583 & 0.436 & 0.655 & 0.539 \\
OLMo-7B & Qwen-7B & 0.83 & 0.07 & 0.748 & \textbf{0.854} & 0.497 & 0.763 & 0.538 \\
\midrule
\textit{Mean} & & & & \textbf{0.663} & 0.618 & 0.463 & 0.623 & 0.560 \\
\bottomrule
\end{tabular}
\end{table}

\begin{table}[h]
\centering
\small
\setlength{\tabcolsep}{4pt}
\caption{GSM8K routing performance (APGR): Full chain-of-thought CMP vs.\ final-answer-only CMP vs.\ P(True). Pairs with gap $<$ 5pp excluded. \textbf{Bold} = best among CMP-Full, CMP-Final, P(True).}
\label{tab:gsm8k-final-apgr}
\begin{tabular}{llcc | ccc | cc}
\toprule
& & & & \multicolumn{3}{c|}{APGR $\uparrow$} & \multicolumn{2}{c}{Baselines} \\
Generator & Verifier & Acc$_g$ & Gap & CMP-Full & CMP-Final & P(True) & G-Ent & G-PPL \\
\midrule
Qwen-0.5B & Qwen-7B & 0.25 & 0.64 & \textbf{0.528} & 0.484 & 0.499 & 0.520 & 0.521 \\
Llama-1B & Qwen-7B & 0.38 & 0.51 & 0.574 & \textbf{0.621} & 0.481 & 0.511 & 0.503 \\
Mistral-7B & Qwen-7B & 0.51 & 0.39 & \textbf{0.590} & 0.524 & 0.506 & 0.575 & 0.513 \\
Qwen-0.5B & Llama-3-8B & 0.25 & 0.36 & \textbf{0.530} & 0.509 & 0.492 & 0.544 & 0.545 \\
Llama-3-8B & Qwen-7B & 0.61 & 0.28 & \textbf{0.614} & 0.579 & 0.488 & 0.551 & 0.538 \\
Llama-1B & Llama-3-8B & 0.38 & 0.23 & \textbf{0.662} & 0.636 & 0.463 & 0.521 & 0.505 \\
Mistral-7B & Llama-3-8B & 0.51 & 0.11 & \textbf{0.998} & 0.967 & 0.483 & 0.872 & 0.638 \\
OLMo-7B & Qwen-7B & 0.83 & 0.07 & 0.956 & \textbf{1.000} & 0.525 & 0.960 & 0.657 \\
\midrule
\textit{Mean} & & & & \textbf{0.682} & 0.665 & 0.492 & 0.632 & 0.552 \\
\bottomrule
\end{tabular}
\end{table}

\end{document}